%% file: main_usenix.tex
\input{macros}

\documentclass[letterpaper,twocolumn,10pt]{article}
\usepackage{usenix2019_v3}

\usepackage{tikz}
\usepackage{amsmath}

\usepackage{filecontents}

\usepackage{multirow}
\usepackage{multicol}
\usepackage{enumitem}
\usepackage{algorithm2e}
\usepackage{algorithmic}
\usepackage{subcaption}
\usepackage{titling}

\usepackage{url}

\usepackage{breakurl}
\usepackage{hyperref}
\makeatletter
\renewcommand{\sectionautorefname}{\S\@gobble}
\renewcommand{\subsectionautorefname}{\S\@gobble}
\renewcommand{\subsubsectionautorefname}{\S\@gobble}
\makeatother
\hypersetup{
    colorlinks,
    linkcolor={blue!50!black},
    citecolor={red!50!black},
    urlcolor={blue!80!black}
}

\usepackage{ifthen}
\newboolean{publicversion}
\setboolean{publicversion}{true}

\ifthenelse{\boolean{publicversion}}{
	\newcommand{\grumbler}[3]{}
        \newcommand{\jm}[1]{}
        \newcommand{\ap}[1]{}
        \newcommand{\nk}[1]{}
        \newcommand{\rr}[1]{}
        \newcommand{\amey}[1]{}
}
{
\newcommand{\grumbler}[3]{\xspace\textcolor{#3}{\bf #1: #2}}
\newcommand{\jm}[1]{{#1}{magenta}}
\newcommand{\ap}[1]{{#1}{violet}}
\newcommand{\nk}[1]{{#1}{teal}}
\newcommand{\rr}[1]{{#1}{cyan}}

\newcommand{\amey}[1]{{#1}{green}}

}

\definecolor{dgreen}{rgb}{0.0, 0.5, 0.0}
\usepackage{comment}
\usepackage{xspace}
\usepackage{authblk}
\usepackage{soul}
\usepackage{tikz}
\usepackage{xcolor}
\usepackage{array}
\usepackage{cleveref}
\usepackage{xparse}

\usepackage{booktabs}
\usepackage{cleveref}%
\crefname{section}{\S}{\SS}
\crefformat{section}{\S#2#1#3} %
\crefformat{subsection}{\S#2#1#3}
\crefformat{subsubsection}{\S#2#1#3}

\crefformat{subfigure}{#2\onlyletter{#1}#3}
\crefrangeformat{subfigure}{(#3\onlyletter{#1}#4--#5\onlyletter{#2}#6)}
\crefmultiformat{subfigure}
  {(#2\onlyletter{#1}#3}
  {,~#2\onlyletter{#1}#3)}
  {,~#2\onlyletter{#1}#3)}
  {,~#2\onlyletter{#1}#3)}

\ExplSyntaxOn
\NewDocumentCommand{\onlyletter}{m}
 {
  \tl_set:Nx \l_tmpa_tl { #1 }
  \tl_item:Nn \l_tmpa_tl { -1 }
 }
\ExplSyntaxOff

\date{}

\title{\Large \bf \sysname: Efficient LLM Inference by Piggybacking Decodes with Chunked Prefills}
\begin{document}
\author[2]{Amey Agrawal\thanks{Work done as intern at Microsoft Research India}}\author[1]{Ashish Panwar} \author[1]{Jayashree Mohan}\author[1]{Nipun Kwatra}\author[1]{Bhargav S. Gulavani}\author[1]{Ramachandran Ramjee} \affil[1]{Microsoft Research India}\affil[2]{Georgia Institute of Technology}

\maketitle

\input{0-abstract}

\input{1-intro}
\input{2-background}

\input{3-motivation}

\input{4-design}

\input{5-eval}

\input{6-discussion}

\input{7-related}

\input{8-conc}

\phantomsection
\label{EndOfPaper}
\clearpage
\bibliographystyle{plain}
\bibliography{all}
\end{document}

%% file: macros.tex
\newcommand{\sysname}{\textsc{Sarathi}\xspace}
\newcommand{\chunking}{\textit{chunked-prefills}\xspace}
\newcommand{\Chunking}{\textit{Chunked-prefills}\xspace}
\newcommand{\hbatch}{\textit{decode-maximal batching}\xspace}
\newcommand{\Hbatch}{\textit{Decode-maximal batching}\xspace}
\newcommand{\hbatches}{\textit{decode-maximal batches}\xspace}
\newcommand{\mycaption}[2]{\caption{\textbf{#1} {#2}}}
\newcommand{\myx}{$\times$\xspace}

\newcommand{\vheading}[1]{\vspace{0.05in}\noindent\textbf{#1}}
\newcommand{\sref}[1]{\S\ref{#1}}

\newcommand{\attnpreproj}{\textit{preproj}\xspace}
\newcommand{\attn}{\textit{attn}\xspace}
\newcommand{\attnpostproj}{\textit{postproj}\xspace}
\newcommand{\ffnlnfirst}{\textit{ffn\_ln1}\xspace}
\newcommand{\ffnlnsecond}{\textit{ffn\_ln2}\xspace}

%% file: 0-abstract.tex
\begin{abstract}

Large Language Model (LLM) inference consists of two distinct phases -- \textit{prefill} phase which processes the input prompt and  \textit{decode} phase which generates output tokens autoregressively. While the prefill phase effectively saturates GPU compute at small batch sizes, the decode phase results in low compute utilization as it generates one token at a time per request.
The varying prefill and decode times also lead to imbalance across micro-batches when using pipeline-parallelism, resulting in further inefficiency due to bubbles.

We present \sysname to address these challenges. \sysname employs \chunking, 
which splits a prefill request into equal sized chunks, and \hbatch, which constructs a batch using a single prefill chunk and populates the remaining slots with decodes. %
During inference, the prefill chunk saturates GPU compute, while the decode requests `piggyback' and cost up to an order of magnitude less compared to a decode-only batch. \textit{Chunked-prefills} allows constructing multiple \hbatches from a single prefill request, maximizing coverage of decodes that can piggyback. Furthermore, the uniform compute design of these batches ameliorates the imbalance between micro-batches, significantly reducing pipeline bubbles.

Our techniques yield significant improvements in inference performance across models and hardware. For the LLaMA-13B model on A6000 GPU, \sysname improves decode throughput by up to 10\myx, and accelerates end-to-end throughput by up to  1.33\myx. For LLaMa-33B on A100 GPU, we achieve 1.25\myx higher end-to-end-throughput and up to 4.25\myx higher decode throughput. When used with pipeline parallelism on GPT-3, \sysname reduces bubbles by 6.29\myx, resulting in an end-to-end throughput improvement of 1.91\myx.

\end{abstract}

%% file: 1-intro.tex
\input{fig-tex/fig-intro}

\section{Introduction}\label{sec:Introduction}

The scaling up of language models~\cite{gpt3-brown2020language,chowdhery2022Palm,openai2022gpt4techreport,kaplan2020scalinglaws} has led to an emergence in their abilities~\cite{wei2022emergent} in a variety of complex tasks --- natural language processing, question answering, code generation, etc. This has led to an explosion in their usage across applications spanning conversational engines~\cite{openai2022gpt4techreport,chatgpt,claudeai,characterai}, search~\cite{bingai,komoai,youdotcom,perplexityai,bard}, code assistants~\cite{githubcopilot,replitghostwriter,amazoncodewhisperer}, etc. The significant GPU compute required for inference on these large models, coupled with their widespread usage, has made LLM inference the dominant GPU workload. Optimizing LLM inference has thus become very important and has seen significant interest recently~\cite{efficiently-scaling-transformer-inference,flexgen,orca}.

In this paper, we first analyze a fundamental reason behind the low efficiency of LLM inference. Each LLM inference request goes through two phases -- a \textit{prefill} phase corresponding to the processing of the input prompt and a \textit{decode} phase which corresponds to the autoregressive token generation. The prefill phase processes \textit{all} tokens in the input sequence in parallel, leading to high GPU utilization even with a small batch size. For example, on an A6000 GPU, for the LLaMA-13B model, a prefill with a sequence length of 512 tokens saturates GPU compute even at a batch size of just one. The decode phase, on the other hand, processes only a \textit{single} token in each autoregressive pass, resulting in very low GPU utilization at low batch sizes. For example, our experiments reveal that, at small batch sizes, the decode cost per token can be as high as $\sim 200$ times the prefill cost per token. Moreover, since a request goes through only a single prefill pass, but multiple decode passes (one for each generated token), the overall inference efficiency is significantly impacted.

One strategy to improve LLM decode efficiency is to increase batch size using model parallelism. In servers with high bandwidth connectivity such as NVIDIA DGX A100, tensor-parallelism~\cite{megatron} can enable deployment of an LLM on up to 8 GPUs, thereby supporting large batch sizes and efficient decode. Pope et al. ~\cite{efficiently-scaling-transformer-inference} show that tensor parallelism can be scaled up to 256 devices on specialized TPUv4 pods. However, tensor-parallelism at such a large scale can result in poor performance when hyper-clusters are unavailable.  In such cases, pipeline parallelism~\cite{pipedream, varuna} can help increase batch size. Thus, systems like Orca~\cite{orca} rely on pipeline parallelism to scale LLM inference and adopt the well-known solution of using micro-batches to mitigate pipeline stalls or bubbles~\cite{gpipe}. However, as we show in this paper, {\it the standard micro-batch-based scheduling can still lead to pipeline bubbles due to the unique characteristics of LLM inference}. Specifically, LLM inference consists of a mixture of varying length prefills and decodes. This creates varying processing times for the different micro-batches, resulting in significant bubbles and wasted GPU-cycles as illustrated in \autoref{fig-intro}(a). Note that the first bubble in the figure is due to varying prompt sizes while the second bubble is due to mismatch between prompt and decode compute times.

In this paper, we present the design and implementation of \sysname, an efficient LLM inference technique. \sysname uses \chunking and \hbatch to address the problems of 1) inefficient decodes and 2) pipeline bubbles. \Chunking splits a prefill request into equal compute-sized chunks. Further, \sysname uses \hbatch to construct a batch by using a single prefill chunk and filling the remaining batch with decodes. This hybrid batch provides units of work that are both compute saturating and uniform, thereby addressing the problems of inefficient decodes and pipeline bubbles.

Since \textit{prefill} and \textit{decode} phases have different compute requirements, the key insight of our method is that mixing prefill and decode requests in a single batch can enable uniformly high compute utilization.  However, since each request has only a single prefill phase, followed by multiple decode phases (for each generated token), we will not have enough prefill requests to be able to always create a hybrid batch of prefills and decodes. \Chunking allows us to construct multiple hybrid batches from a single prefill request, thereby increasing the coverage of decodes that can piggyback with a prefill. In our hybrid batch, the single prefill chunk ensures high GPU utilization, while the decode phase requests `piggyback' along. Given an average prefill-to-decode token ratio for an LLM application, we select a prefill chunk size that maximizes the overall performance.

The hybrid batches constructed in \sysname have a \textit{uniform} compute requirement. Thus, when used with pipeline parallelism, \sysname ensures that the micro-batches are well balanced, which results in a significant reduction in pipeline bubbles as shown in \autoref{fig-intro}(b).

We evaluate \sysname across different models and hardware --- LLaMA-13B on A6000 GPU, LLaMA-33B on A100 GPU, and GPT-3 with 8-way pipeline and 8-way tensor parallelism across a simulated cluster of 64 A100 GPUs. For LLaMA-13B on A6000, \sysname improves decode throughput by up to 10\myx and results in up to 1.33\myx end-to-end throughput improvement. Similarly, for LLaMA-33B on A100, our decode throughput improves by 4.25\myx, and results in a 1.25\myx end-to-end throughput improvement. When used with pipeline parallelism, \sysname reduces bubbles by 6.29\myx, resulting in end-to-end speedup of 1.91\myx.

The main contributions of our paper include:
\begin{enumerate}[noitemsep,topsep=0pt,parsep=0pt,partopsep=0pt]
    \item \Chunking which allows the construction of work units that are compute saturating and uniform.
    \item \Hbatch which allows inefficient decodes to `piggyback' with efficient prefills.
    \item Application of \chunking and \hbatch to pipeline parallelism to significantly reduce pipeline bubbles.
    \item Extensive evaluation over multiple models, hardware, and parallelism strategies demonstrating up to 1.91\myx improvement in throughput.
\end{enumerate}

%% file: fig-tex/fig-intro.tex
\begin{figure}[!t]
    \centering
    \includegraphics[width=0.48\textwidth]{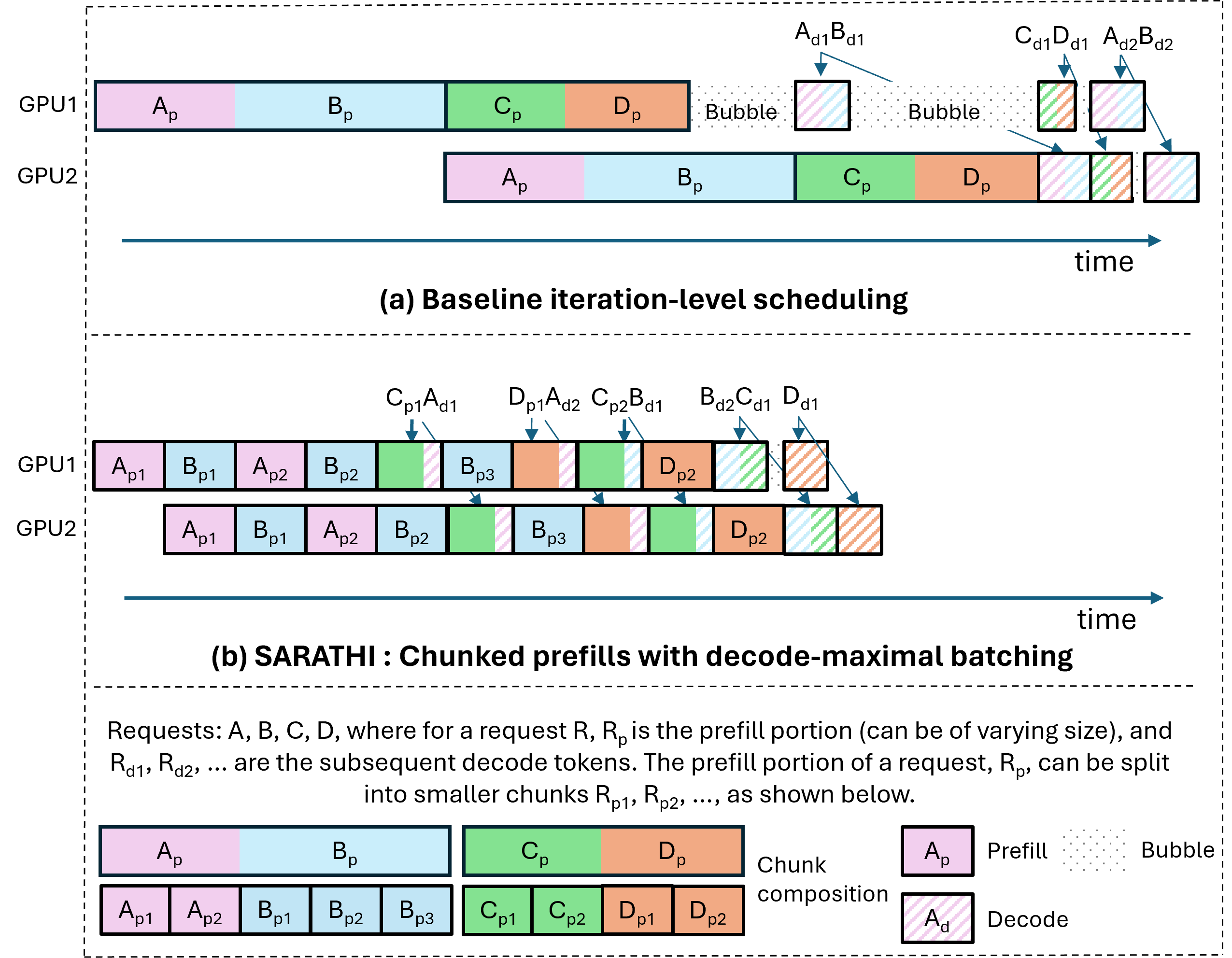}
    \mycaption{Example two-stage pipeline parallel schedule.}{(a) In prior solutions like Orca~\cite{orca}, pipeline bubbles are common due to varying prompt and decode compute times. Further, decodes are highly inefficient (decode {\it cost-per-token} is order-of-magnitude higher than Prefill). (b) \sysname significantly reduces pipeline bubbles and enables more efficient {\it piggybacked decodes}.}
    \label{fig-intro}
\end{figure}

%% file: 2-background.tex
\section{Background}
\label{sec:background}

We first give an overview of the transformer architecture, followed by a brief discussion of the two phases of LLM inference, and pipeline parallelism.

\subsection{The Transformer architecture}
\label{sec:background:arch}
\autoref{fig:transformer:architecture} shows the architecture of a transformer decoder block. Each decoder block consists of two primary modules: self-attention and feed-forward network (FFN). These modules can be divided into the following six operations:
\attnpreproj, \attn, \attnpostproj (within the attention module), and \ffnlnfirst, \ffnlnsecond (within FFN) and {\it others} (e.g., layer normalization, activation functions, residual connections etc.). 

\input{fig-tex/fig-transformer-arch}

\subsection{The prefill and decode phases}
\label{sec:background:prefill:decode}

Transformer inference begins with the prefill phase that processes all the input tokens of a given batch in parallel. In this phase, the input to a transformer block is a tensor $X$ of shape $[B,L,H]$ where $B$ denotes the batch size, $L$ denotes the sequence length of each request (i.e., the number of input tokens in the given query), and $H$ is the model's embedding size (e.g., 5120 for LLaMA-13B). 

\autoref{table:tensor:shapes} shows the shapes of input, output, and weight tensors of the various operations. Each transformer block first computes self-attention on a given input $X$.  Typically, multi-head attention is used, but we consider only one head for simplicity of exposition. A linear transformation \attnpreproj over X (using the weight tensors $W^Q$, $W^K$ and $W^V$ of shape $[H,H]$) produces the $Q$, $K$ and $V$ that are commonly known as queries, keys, and values, each of shape $[B,L,H]$. Internally, \attnpreproj is a single matrix-matrix multiplication of $X$ with a combined weight tensor of shape $[H,3H]$.

Next, the \attn computation over $Q$, $K$ and $V$ produces a tensor $Y$ of shape $[B,L,H]$. Finally, \attnpostproj applies a linear transformation over $Y$ (using weight matrix $W_o$ of shape $[H,H]$), returning a tensor $Z$ of shape $[B,L,H]$.

Next, the FFN module performs two batched matrix-matrix multiplications. In \ffnlnfirst, $Z$ is multiplied with a weight tensor of shape $[H,H_2]$ producing an output tensor of shape $[B,L,H_2]$, which is then multiplied by a weight tensor of shape $[H_2,H]$ in \ffnlnsecond to output a tensor of shape $[B,L,H]$. Here, $H_2$ refers to the second hidden dimension of the model.%

The decode phase performs the same operations as prefill, but only for the \textit{single} token which was generated in the last autoregressive iteration. Thus, the input tensor in decode phase is of shape $[B,1,H]$ (as opposed to $[B,L,H]$ of prefill). Further, the attention computation for each new token depends on the key ($K$), and value ($V$) tensors of all prior tokens in the same request. To avoid recomputing $K$ and $V$ of all tokens in every iteration, most implementations cache these values in GPU memory - which is referred to as the KV cache. Note that each token's $K$ and $V$ tensors are of shape $[1,H]$.

\input{tables/tbl-tensor-shapes}
\subsection{Multi-GPU LLM Inference}
\label{sec:background_multi_GPU_inference}
As the model sizes of LLMs increase, it becomes necessary to scale them to multi-GPU as well as multi-node deployments~\cite{efficiently-scaling-transformer-inference,multinodeinferenceblog}. Furthermore, LLM inference throughput, specifically that of the decode phase is limited by the maximum batch size we can fit on a GPU. Inference efficiency can therefore benefit from model-parallelism which shards the model weights across multiple GPUs freeing up memory to support larger batch sizes. Prior work has employed both tensor-parallelism (TP)~\cite{megatron} (within node) and pipeline-parallelism (PP)~\cite{orca, fastertransformer, fastserve} (across nodes) for this purpose.

TP shards each layer across the participating GPUs. This splits both the model weights and KV cache equally across GPU workers, leading to linear scaling of per-GPU batch size. However, it comes at a high communication cost due to two all-reduce operations per layer -- one in attention computation and the other in FFN~\cite{megatron}. Moreover, since these communication operations are in the critical path, TP is preferred only within a single node connected by high bandwidth interconnects like NVLink. PP is primarily used to facilitate cross-node deployments for very large models, where the model cannot fit within a single node.

Compared to TP, PP splits a model layer-wise, where each GPU is responsible for a subset of layers. To keep all GPUs in the `pipeline' busy, \textit{micro-batching} is employed. These micro-batches move along the pipeline from one stage to the next at each iteration. PP has the advantage of a much better compute-communication ratio compared to TP, as we only need to send activations once for multiple layers of compute. Furthermore, PP requires communication only via point-to-point communication operation, compared to the more expensive allreduces required in TP. Thus, PP is the only viable model-parallelism approach when high-bandwidth connectivity like NVlink is unavailable at cluster-scale. In such settings, the use of PP can help increase the maximum batch size supported in each node by 2-3\myx, thereby improving LLM inference efficiency.

%% file: fig-tex/fig-transformer-arch.tex
\begin{figure}
\center
\includegraphics[width=\columnwidth]{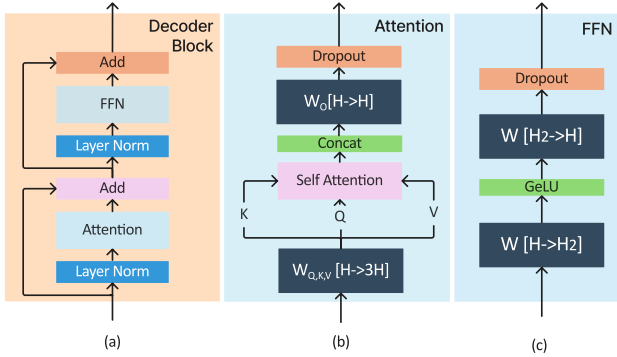}
\caption{High-level architecture of a decoder block.}
\label{fig:transformer:architecture}
\end{figure}

%% file: tables/tbl-tensor-shapes.tex
\begin{table}[t]
\center
\scalebox{0.9}{
\begin{tabular}{l|rrr}
\multirow{2}{*}{Operation} & \multicolumn{3}{c}{Shapes of tensors} \\ \cline{2-4}
 & Input(s) & Weight(s) & Output(s) \\ \hline
\attnpreproj & $[B,L,H]$ & $[H,H]$ & $[B,L,H]$ \\
\attn & $[B,L,H]$ & - & $[B,L,H]$ \\
\attnpostproj & $[B,L,H]$ & $[H,H]$ & $[B,L,H]$ \\
\ffnlnfirst & $[B,L,H]$ & $[H,H_2]$ & $[B,L,H_2]$ \\
\ffnlnsecond & $[B,L,H_2]$ & $[H_2,H]$ & $[B,L,H]$ \\ \hline
\end{tabular}}
\caption{Shapes of the input, weight, and output tensors in a transformer decoder block. B, L and H denote batch size, embedding (aka hidden) size and sequence length (L=1 during decode, except for attention).}
\label{table:tensor:shapes}
\end{table}

%% file: 3-motivation.tex
\section{Motivation}
\label{sec:motivation}

In this section, we show that LLM inference is inefficient for two main reasons: (1) the decoding phase is memory-bound, and (2) the use of pipeline parallelism leads to significant pipeline bubbles for LLMs. Together, these factors lead to poor GPU utilization for LLM inference.

\subsection{Analyzing Prefill and Decode Throughput}
\label{sec:analysis:throughput}

\input{fig-tex/fig-analysis-op-breakdown}

\autoref{fig:analysis_per_token} shows the per-token cost of each of the six transformer operations (\autoref{sec:background:arch}) for prefill and decode at various batch sizes for a fixed sequence length (prefill+decode) of 1024. First, we observe that prefill has almost constant per-token cost across various batch sizes, indicating that prefill saturates the GPU even at batch size of 1. Second, we see that decode behaves very differently from prefill as the per-token cost reduces significantly when the batch size increases. Third, we see that the  
{\it decode cost per-token is 200\myx, 100\myx, and 16.7\myx that of prefill at batch size of 1, 2 and 18, respectively}. Thus, it is clear that optimizing decodes is critical for efficient LLM inference. Finally, we see that the operations under {\it others} contribute less than 5\% of the overall runtime of the transformer block.  Hence, we focus on only optimizing the five major operations and ignore others.

\input{fig-tex/fig-mot-ai}
\autoref{fig:mot_tput} shows the throughput of the prefill and decode stages for different batch sizes ($B$) and sequence lengths ($L$). We observe that the throughput of the prefill phase saturates at about 180 tokens/millisecond when $B\times L\geq 512$: e.g., a single prefill request can achieve peak throughput at $L\geq 512$. In contrast, the decode throughput increases linearly with small batch sizes. To further understand the saturation point of decode phase, we profile a single layer as opposed to the 40 layers of the full model. This enables us to fit 40\myx larger batches on the GPU due to the reduced memory footprint of model weights and KV caches. We find that decode saturates at a much larger batch (e.g., 256 with 1024 sequence length).  Such large batches are infeasible to run with the full model.

To explain this behavior, we profile the arithmetic intensity of individual operations: arithmetic intensity captures the amount of compute per memory read/write that can be used to distinguish between compute-bound and memory-bound operations.~\autoref{fig:mot_ai} shows the arithmetic intensity of each operation separately for prefill (left) and decode phases (right). As shown, in prefill phase, all operations have high arithmetic intensity, even at a batch size of {\em one}. On the other hand, the arithmetic intensity of these operations drop by more than two orders of magnitude in the decode phase. Only at a very large batch size of 256, the decode phase starts becoming compute-intensive. However, scaling up the batch size to such high values is infeasible due to the KV-cache footprint of each request. For instance, we can fit a maximum batch size of 18 requests at a sequence length of 1K for the LLaMA-13B model on an A6000 GPU. Therefore, in the range of batch sizes that are practical today, the decode phase remains memory-bound.

The difference between the throughput scaling of these two phases stems from the fact that the prefill phase computes (batched) {\em matrix-matrix multiplications} as opposed to the {\em vector-matrix multiplications} of the decode phase. It is well-known that kernels with arithmetic intensity above a GPU's FLOPS:MemBandwidth ratio are compute-bound and can be executed efficiently~\cite{tile-quantization}. In contrast, kernels with a lower arithmetic intensity fail to utilize GPUs well due to being memory-bound.

\subsection{Pipeline Bubbles in LLM Inference}
Pipeline Parallelism (PP) is a popular strategy for cross-node deployment of large models, owing to its lower communication overheads compared to Tensor Parallelism (TP). PP splits a model layer-wise, where each GPU is responsible for a subset of layers; compared to TP which shards each layer across the participating GPUs. As discussed in ~\autoref{sec:background_multi_GPU_inference}, compared to TP, PP has a much better compute-communication ratio and does not require expensive interconnects.

A challenge with PP, however, is that it introduces \textit{pipeline bubbles} or periods of GPU inactivity as subsequent pipeline stages have to wait for the completion of the corresponding micro-batch in the prior stages. Pipeline bubbles is a known problem in training jobs, where they arise between the forward and backward passes due to prior stages needing to wait for the backward pass to arrive. Micro-batching is thus commonly employed in PP training jobs to amortize the bubbles across the multiple micro-batches forming a batch~\cite{varuna,pipedream,gpipe}. 

Unlike training, since inference jobs only do forward passes and do not have backward passes, one might expect that the use of micro-batches will fully avoid pipeline bubbles during inference. In fact, prior work on transformer inference, such as, FasterTransformer~\cite{fastertransformer} and FastServe~\cite{fastserve} use micro-batches and do not consider the problem of bubbles with PP.  

Orca~\cite{orca} suggests that the use of iteration-level scheduling eliminates bubbles in pipeline scheduling (see Figure 8 in ~\cite{orca}). 
However, as we show in this paper, even with iteration-level scheduling of requests, each micro-batch (or iteration) in LLM inference can require a different amount of compute (and consequently has varying execution time), depending on the composition of prefill and decode tokens in the micro-batch (see ~\autoref{fig-mot-bubbles}). We identify three types of bubbles during inference: (1) bubbles like $PB_1$ that occur due to the varying number of prefill tokens in two consecutive micro-batches (2) bubbles like $PB_2$ that occur due to different compute times of prefill and decode stages when one is followed by the other, and (3) bubbles like $PB_3$ that occur due to difference in decode compute times between micro-batches since the accumulated context length (KV cache length) varies across requests. These pipeline bubbles are wasted GPU cycles and directly correspond to a loss in serving throughput with pipeline parallelism. If we can ensure that each micro-batch performs uniform computation, we can mitigate these pipeline bubbles. 

\input{fig-tex/fig-mot-bubbles}

\subsection{Insights}
\label{sec-mot-insights}

Our experiments show that the prefill and decode stages have very different compute utilization patterns -- prefill can saturate GPU compute even with a single request, while decodes require a large batch size to be compute-efficient. However large batches are impractical due to their high KV cache footprint. Such disproportionate resource utilization implies that for every request, there are phases of high compute utilization due to efficient prefills, followed by a potentially long tail of inefficient decodes which results in poor overall GPU utilization. Furthermore, the non-uniformity in compute times across micro-batches leads to pipeline bubbles, resulting in inefficient pipeline parallel multi-GPU deployments.

This observation leads us to our key insight that it is possible to construct uniformly compute-intensive batches by (1) slicing a large prefill request into smaller compute-efficient and uniform chunks using \chunking and (2) creating a hybrid batch of a prefill chunk and piggybacking decodes alongside this chunk. Consequently, creating such uniform and compute-intensive batches ensures high GPU utilization throughout, as well as, minimizes pipeline bubbles in multi-GPU deployments by eliminating the runtime variance across micro-batches in different stages of the pipeline.

%% file: fig-tex/fig-analysis-op-breakdown.tex
\begin{figure}[t]
    \centering
    \includegraphics[scale=0.25]{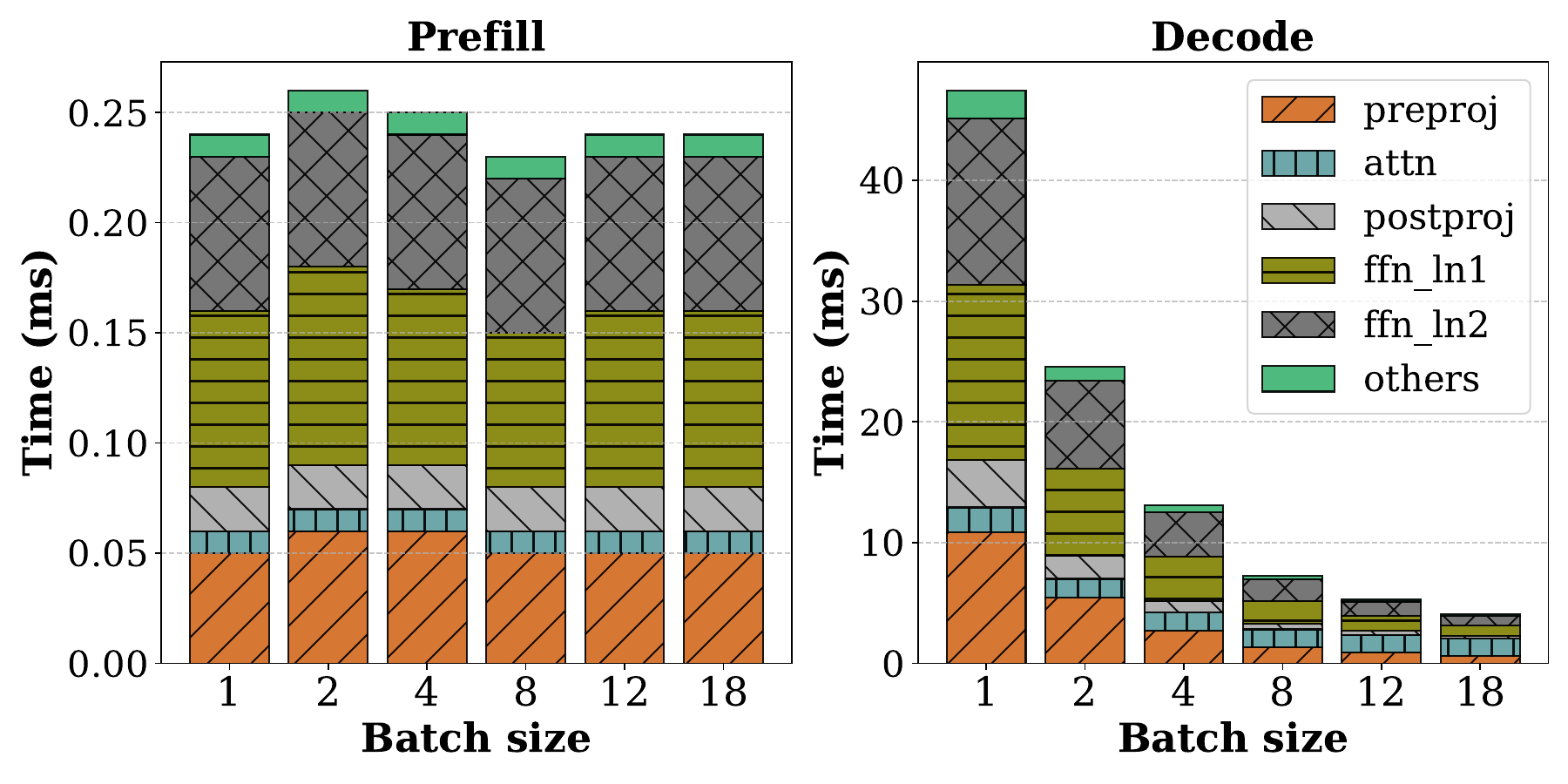}
    \caption{Per-token prefill and decode time with different batch sizes (sequence length = 1024) for LLaMa-13B on A6000 GPU. Prefill saturates GPU compute even at batch size of 1 and results in almost constant per-token time across batch sizes. Decode under-utilizes GPU compute and costs as much as 200\myx prefill for batch size 1. The incremental cost of linear operators for decode is almost zero as batch size increases. The attention cost does not benefit from batch size as it is memory-bound.}
    \label{fig:analysis_per_token}
\end{figure}

%% file: fig-tex/fig-mot-ai.tex
\begin{figure}[!t]
    \centering
        \begin{subfigure}[b]{0.48\textwidth}
        \includegraphics[width=\textwidth]{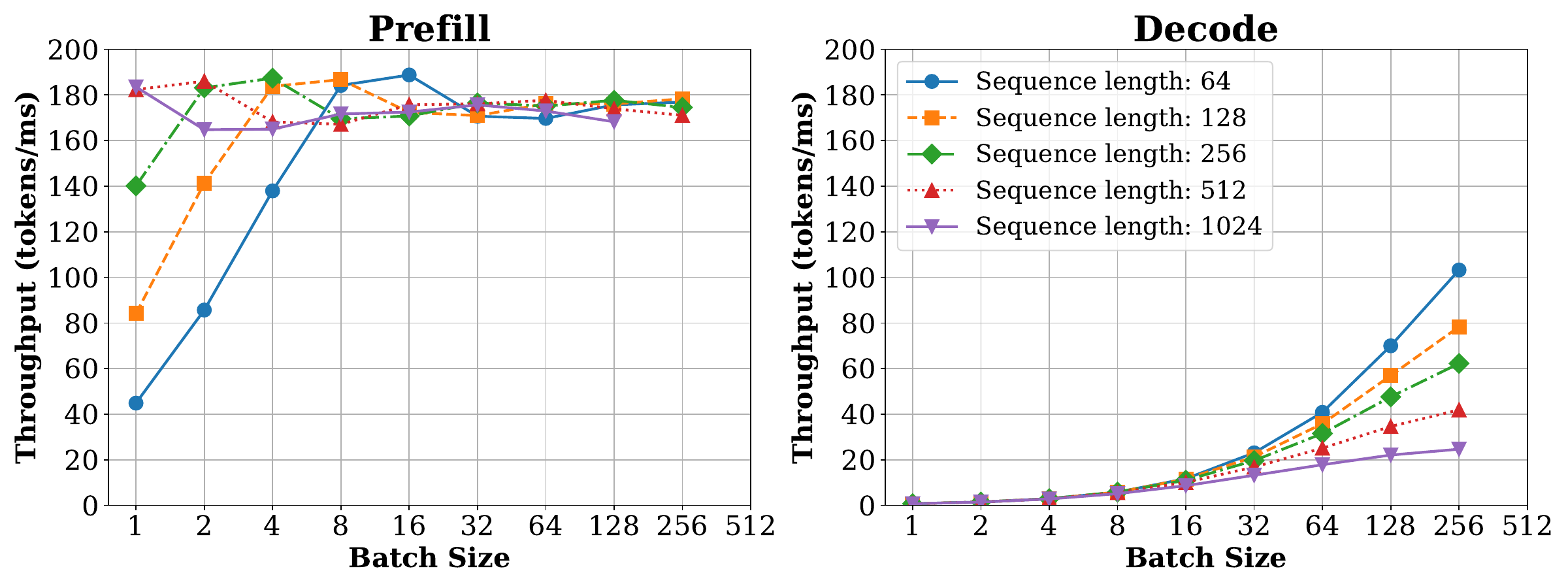}
        \caption{Throughput of a single layer of LLaMA-13B on A6000 GPU.}
        \label{fig:mot_tput}
    \end{subfigure}
    \begin{subfigure}[b]{0.48\textwidth}
        \includegraphics[width=\textwidth]{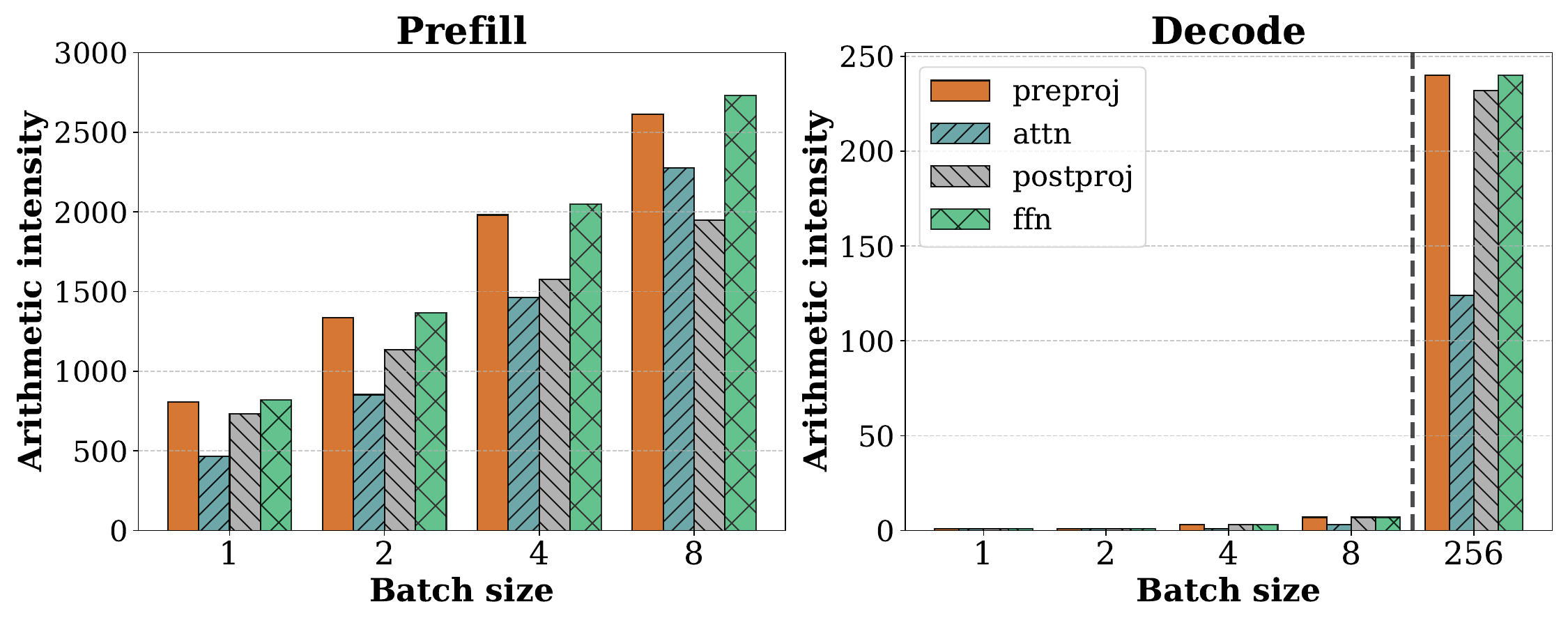}
        \caption{Arithmetic intensity with 1K sequence length (per-request).}
        \label{fig:mot_ai}
    \end{subfigure}
    \caption{Impact of the arithmetic intensity (bottom) on the throughput (top) of prefills and decodes for LLaMA-13B on A6000 GPU.}
    \label{fig:throughput_and_ai}
\end{figure}

%% file: fig-tex/fig-mot-bubbles.tex
\begin{figure}
    \centering
    \includegraphics[width=0.5\textwidth]{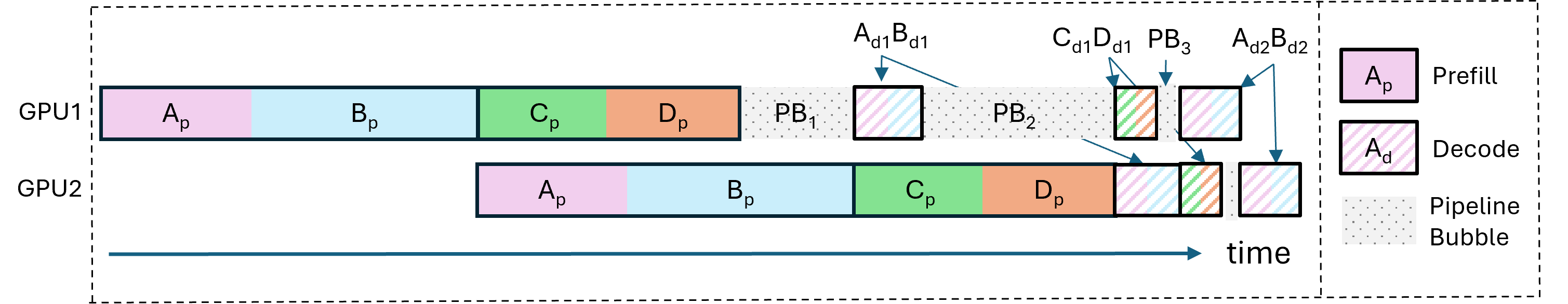}
    \mycaption{Pipeline bubbles in LLM inference}{A 2-way PP iteration-level schedule~\cite{orca} across 4 requests (A,B,C,D) shows the existence of pipeline bubbles due to non-uniform batch execution times.}
    \vspace{-1em}
    \label{fig-mot-bubbles}
\end{figure}

%% file: 4-design.tex
\section{\sysname : Design and Implementation}\label{sec:design}

In this section, we describe the design and implementation of \sysname, which employs two techniques - \chunking and \hbatch to improve the performance of LLM inference.
\subsection{Overview}
Conventional inference engines like FasterTransformer~\cite{fastertransformer} perform request-level inference scheduling. They process batches at request granularity; i.e., the they pick the next batch of requests to execute on the model replica only when all the requests in the current batch complete. While this reduces the operational complexity of the scheduling framework, it is inefficient in its use of resources. Shorter requests in a batch have to be padded to match the length of the longest request, and thus does wasteful work instead of exiting early. Alternatively, iteration-level scheduling has been proposed in more recent systems like Orca~\cite{orca}, vLLM~\cite{vLLM:github}, and HuggingFace TGI ~\cite{hftgi}, where depending on the predetermined batch size $b$, requests can dynamically enter and exit a batch.

However, today's iteration-level scheduling systems do not pay attention to the requests that comprise the batch, and the varying execution time between batches. Specifically, a batch could comprise of requests only in the prefill phase, requests only in the decode phase, or mixed requests consisting of a few prefills and decodes, with the only constraint that the batch size is $b$ at all times. As discussed in ~\sref{sec-mot-insights}, such batch formation results in non-uniform units of compute, resulting in periods of bursty resource utilization, and pipeline bubbles. \sysname tackles this challenge by introducing two key techniques: \chunking and \hbatch.

\subsection{\Chunking}
\label{sec-design-chunkedprefill}

\Chunking is a prefill splitting mechanism hinged on two key insights. First, for a given model and GPU, increasing the number of prefill tokens shows diminishing returns in throughput beyond a certain point as shown in \autoref{fig:mot_tput}. For instance, the Llama-13B model achieves peak prefill throughput on an A6000 GPU when the number of prefill tokens is 512 or higher. At a chunk size of 256, we see a marginal reduction of 12.5\% in the peak throughput. Further, as the size of the hidden dimension in the model increases, the chunk size needed to saturate the GPU compute drops; for example, the throughput of a single layer of GPT-3 (hidden size = 12288) peaks at a chunk size of 256 on an A100 GPU. This implies that a compute-saturating batch can be formed with a carefully sliced prefill chunk. Second, in many practical scenarios, the size of prefill is reasonably large, ranging from 1K -- 4K in production workloads, thereby opening the doors for chunking a prefill request into smaller units of compute.

\input{fig-tex/fig-attn-chunk-prefills}

Implementing \chunking requires carefully setting the attention mask. If a request's input prompt of say size 1K is split into four chunks of size 256 tokens each, we need to ensure that the attention masks are appropriately set for every subsequent prefill chunk until the end of the prompt. For ease of exposition, using an example of chunk size of four, \autoref{fig-attn-chunk-prefills} shows how \sysname progressively sets the attention mask for every successive chunk of a prefill prompt in three consecutive iterations: each query token $q_i$ can peek into the keys (and values) of all the tokens preceding it, but not the ones that follow. Setting the attention mask this way ensures that \chunking computation is mathematically equivalent to the full prefill.

\vheading{Overhead of \chunking}. Splicing the input of a prefill sequence into multiple smaller chunks has two potential sources of overhead. First, the arithmetic intensity of \chunking computation decreases as the chunk size becomes smaller. Therefore, smaller chunks can affect prefill efficiency due to low GPU utilization. However, this can be addressed easily with a one-time profiling of the prefill throughput for various chunk sizes on a given model-hardware combination and expected workloads and a chunk size can be chosen such that the end-to-end throughput of the model is maximized.

Second, \chunking pose a slight overhead in attention computation due to repeated memory accesses of the KV cache of a request's tokens from prior chunks. While every \chunking operation until the end of the prompt will perform the same number of computations for FFNs, the attention kernel in every subsequent chunk after the first will have to reread all the KV pairs of the prior tokens from the GPU memory, as shown in \autoref{fig-attn-chunk-prefills}. For example, if a prefill sequence is split into $N$ chunks, then the first chunk's KV cache is loaded $N$ times, the second chunk's KV cache is loaded $N-1$ times, and so on.
However, the overhead due to increased attention time does not significantly affect the end-to-end prefill efficiency because attention computation is a small fraction of the overall forward pass time as seen in Table~\ref{tbl-compute-split}. We present a detailed analysis of the overheads of \chunking in \autoref{sec:eval:ablation}.

\subsection{Decode-Maximal Batching}
\label{sec-design-mixedbatch}
Harnessing the benefits of \chunking requires us to carefully construct a hybrid batch consisting of a mix of prefill and decode tokens, so as to maximize compute utilization and ensure uniform compute time across all batches. We propose \hbatch to alleviate the imbalance in compute and memory utilization in iterative scheduling by exploiting the idea of \chunking. 

In \hbatch, we construct a batch by using a single prefill chunk and piggybacking the remaining slots with decode tokens. This hybrid batch provides us with units of work that are both compute saturating and uniform. We now discuss how we construct a hybrid batch to achieve maximum efficiency.

\subsubsection{Piggybacking decodes with prefills} 
To piggyback decodes with a prefill, we need to take care of two things. First, we need to identify the maximum possible batch size of decodes that can be piggybacked and also identify the number of prefill tokens that comprise the prefill chunk. Second, in order to actually utilize the GPU-saturating prefill computation of the hybrid batch to make the decodes efficient, we need to \textit{fuse} the linear operation computations for the prefill chunk and decodes of the batch into a single operation.

\vheading{Decode batch}. The maximum decode batch size to be piggybacked with a prefill chunk is determined based on the available GPU memory ($M_G$), the model's parameter memory requirement per GPU ($M_S$), and the maximum sequence length $L$ that the model supports. The total of prefill ($P$) and decode ($D$) tokens per request cannot exceed this maximum sequence length. Assuming the memory required per pair of K and V for a token is $m_{kv}$, %
the maximum permissible batch size $B$ is determined as follows
\begin{equation*}
    B = \lfloor \left(\frac{M_G - M_S}{L*m_{kv}}\right) \rfloor
\end{equation*}

In the baseline scheme, decode-only batches can be of size at most $B$. In \sysname, the number of decodes can be at most $B-1$ as they piggyback along with one prefill chunk (the prefill's KV cache also needs to be in GPU memory until its corresponding decode iterations begin).

In \hbatch, we fuse all the linear operations, while letting the attention computations for the prefill and decodes happen separately. The attention operation for decode requests is batched together, while the attention in prefill chunk is processed separately.

\vheading{Decode efficiency}. Recall that the prefill and decode phases follow the same computation path, i.e., the linear operations use the same weight tensors in both the prefill and decode phases. However, compared to prefill, a decode iteration consists of only a few input tokens (equal to the batch size). Therefore, most of the computation time in baseline decoding is spent fetching model weights from GPU's global memory.

In contrast, \hbatch computes over the decode tokens using matrix matrix multiplications, by combining decode tokens with the prefill tokens in a single matrix multiplication operation. This, effectively eliminates the need to load the model weights separately for decoding --- i.e., once the model weights are fetched for prefills, they are also reused for decoding. As a result, \hbatch converts decoding from being in a memory-bound phase to being in a compute-bound phase. This way, decodes, when piggybacked with prefills come at a marginal cost in \sysname (note that the attention cost remains unchanged).

\input{tables/tbl-compute-split}

To illustrate the various costs involved through an example, \autoref{tbl-compute-split} compares the runtime of one iteration of \hbatch with that of the baseline scheme that computes prefill and decode iterations separately. With baseline batching, a decode-only iteration spends 12.49 milliseconds per token. In contrast, per-token decode time is only 1.2 milliseconds with \hbatch. This shows that piggybacking decodes with prefills can improve decode throughput by up to an order to magnitude. 

\subsection{Identifying the ideal chunk size}

An important design consideration in \sysname is {\em how to pick the most suitable chunk size}. A straightforward choice is to pick the smallest chunk size that saturates a model's prefill throughput. However, we find that this strategy is not the most efficient in many cases.

To demonstrate the importance of chunk size, we introduce a simple notation ``P:D ratio" that is computed as the ratio of the number of prefill tokens to the number of decode tokens in a given batch. For example, a P:D ratio of 10 implies that the number of prefill tokens is 10 times that of decode. For a fixed P+D, a lower value of P:D ratio means that there are more decode tokens in a batch compared to one with a higher value of P:D ratio.

\input{fig-tex/fig-tile-quantize}

The size of prefill chunks in \sysname impacts the number of decodes that can be piggybacked using \hbatch. For example, consider a batch size of four requests (where one request is in the prefill phase and three are in the decode phase) and a chunk size of 128. A prefill of size P will then yield P/128 prefill-chunks, allowing P/128 $\times 3 \approx P/42$ decodes to piggyback. Thus, in this case, when the P:D ratio is greater than 42, it allows us to overlap all decodes with prefills. Similarly, if the chunk size is 256, then all decodes can be piggybacked when the P:D ratio is greater than 84. Therefore, a lower chunk size can help piggyback more decode tokens for a given prefill sequence.

Note that decoding time increases as the the P:D ratio goes down. Therefore, beyond a certain point, optimizing decodes becomes more important than executing prefills at peak efficiency.  For example, if the prefill and decode phases consume 10\% and 90\% of the total time, respectively, then even a $5\times$ overhead in prefills is acceptable if the decodes can be optimized by $2\times$ or more.

To sum it up, identifying a suitable chunk size involves a trade-off: smaller chunks piggyback more decodes but at the expense of lower prefill efficiency whereas larger chunks are prefill efficient but piggyback fewer decodes. Therefore, the ideal chunk size depends on the expected P:D ratio and the split between prefill and decode times for a given application.

\vheading{The tile quantization effect.} Additionally, we observe an intricate detail related to the chunk size. GPUs compute matmuls by partitioning the given matrices into tiles and assigning them to different thread blocks for parallel computation. Here, each thread block refers to a group of threads and computes the same number of arithmetic operations. Therefore, matmuls achieve maximum GPU utilization when the matrix dimensions are divisible by the tile size.  Otherwise, due to {\em tile quantization}, some thread blocks perform extraneous (wasted) computation~\cite{tile-quantization}.

Notice that the time to compute a prefill sequence suddenly increases when the sequence length is just higher than a multiple of 128 (tile size in our experiments). For example, as shown in~\autoref{fig-tile-quantize}, doubling the sequence length from 128 to 256 tokens increases iteration time by 27\% --- from 55ms to 69.8ms. However, adding only a single token further increases the iteration time to 92.33ms --- a dramatic 32\% increase due a only one additional token. This shows that the GPU is most efficient at matmuls when the sequence length is a multiple of the tile size.

Therefore, selecting the ideal chunk size is a two-fold decision. First, pick a chunk size based on the desired prefill efficiency for the given workload. Next, ensure that the sum of chunk size and the number of piggybacked decode tokens is a multiple of the tile size. This ensures that the relevant matrix dimension of the fused operations stays a multiple of the tile size. For example, if the chosen chunk size is 256, the tile size is 128, and the maximum permissible batch size is $B$, then, the prefill chunk size should be $256-(B-1)$.

\input{4.1-impl}

%% file: fig-tex/fig-attn-chunk-prefills.tex
\begin{figure}
    \centering
    \includegraphics[trim=50 415 80 72, clip, scale=0.52]{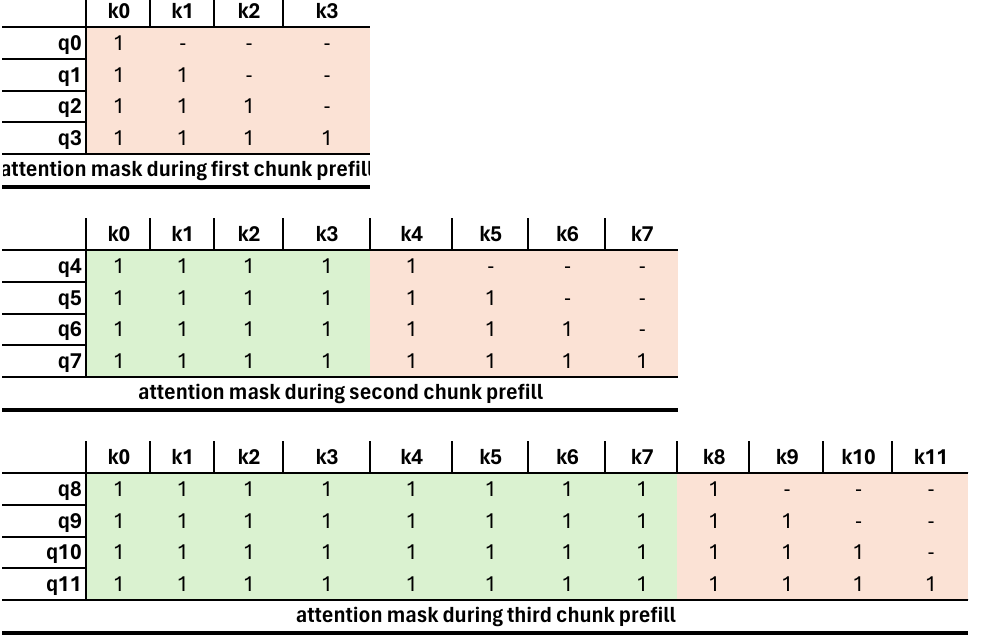}
    \caption{Example of how attention mask is set across different chunk prefill iterations in \sysname (q and k represent ``query" and ``key" tokens, respectively). The attention mask for v (``values") is set similarly.}
    \label{fig-attn-chunk-prefills}
\end{figure}

%% file: tables/tbl-compute-split.tex
\begin{table}[t]
\center
\scalebox{0.9}{
\begin{tabular}{c|cc|c|cc}
\textbf{Batching} & \multicolumn{2}{c|}{\textbf{Operation(s)}} & \textbf{Total} & \multicolumn{2}{c}{\textbf{Per-token Time}} \\ \cline{2-3} \cline{5-6} 
\textbf{Scheme} & Linear & Attn  & \textbf{Time} & Prefill & Decode  \\ \hline
Prefill-only & 224.8 & 10 & 234.8 & 0.229 & - \\
Decode-only & 44.28 & 5.68 & 49.96 & - &  \textcolor{red}{\textbf{12.49}} \\
Decode-maximal & 223.2 & 15.2 &  238.4 & 0.229 & \textcolor{dgreen}{\textbf{1.2}} \\
\bottomrule
\end{tabular}}
\mycaption{Per-token prefill and decode time (in ms)}{For LLaMA-13B on A6000 GPU, the rows show operation times for 1) prefill-only requests of prompt size 1024 of batch size 4, 2) decode-only batch size of 4 with sequence length 1024, and c) a mixed batch of a single 1021 prefills and 3 decodes. \Hbatch reduces the decode time per token by an order of magnitude.}
\label{tbl-compute-split}
\end{table}

%% file: fig-tex/fig-tile-quantize.tex
\begin{figure}
    \centering
    \includegraphics[trim={60 0 0 10}, clip, scale=0.25]{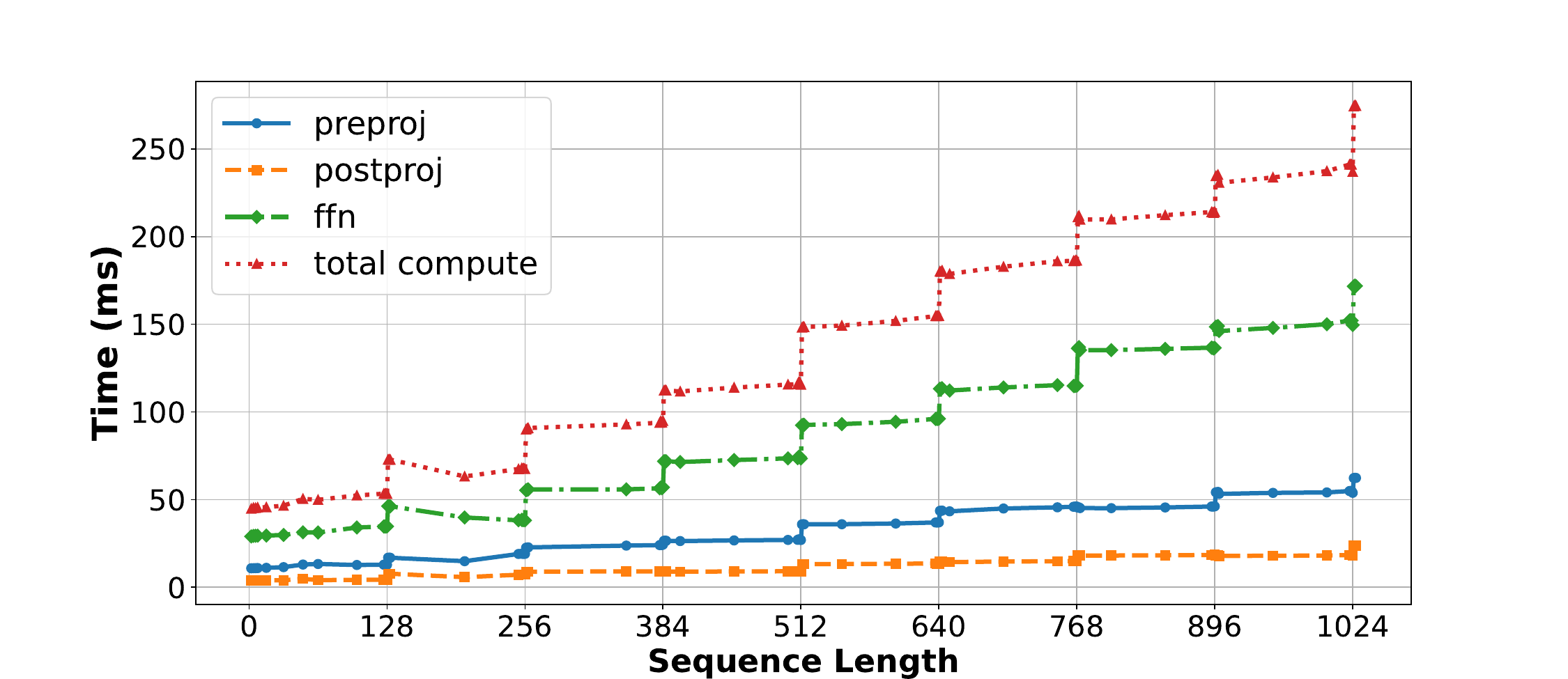}
    \caption{The effect of tile quantization on the runtime of one iteration of LLaMA-13B on A6000 GPU.}
    \label{fig-tile-quantize}
\end{figure}

%% file: 4.1-impl.tex
\subsection{Implementation}
\label{sec-impl}

We implement \sysname on the nanoGPT codebase~\cite{nanogpt} with support for both \chunking and \hbatch. To compare against Orcas's iteration-level scheduling, we use our mixed batching mechanism, with no constraint on the number of prefills allowed per batch. This ensures that there is no discrepancy in results between the baselines and \sysname due to differences in implementation. To compute the attention operation, we use xformers implementation~\cite{xformers} as in our setup, it outperformed PyTorch 2.0's in-built attention implementations: i.e., flash attention, memory-efficient attention, and math attention kernels. To avoid allocating memory for KV caches in each decode iteration, we pre-allocate the KV cache as per the maximum sequence length for each experiment and update respective KV pairs in place when required.

We support different model configurations in our codebase to evaluate \sysname over different model and hardware combinations. For example, to evaluate LLaMA-13B, we set the number of layers and attention heads to 40, and hidden size to 5120. For LLaMA-33B, we use 60 layers, 52 attention heads, and hidden size of 6656. For GPT-3, we use 96 layers, 96 attention heads, and hidden size of 12288. The configurations are as per the publicly available architectural parameters of these models~\cite{llamamodels, gpt3}.

%% file: 5-eval.tex
\section{Evaluation}
\label{sec-eval}

\input{tables/tbl-eval-setup}

We evaluate \sysname on a variety of models and GPUs using physical deployments for single GPU experiments and profile-driven simulations for large-scale experiments as shown in~\autoref{tbl-eval-setup}. 
Our evaluation seeks to answer the following questions:
\begin{enumerate}[leftmargin=*] \itemsep0em 
    \item What is the impact of \sysname on the throughput of decodes as well as the end-to-end throughput of LLMs? In addition, what is the impact of varying sequence lengths, batch sizes, and P:D ratios (\autoref{sec:eval:end-to-end})?
    \item How does \sysname compare to existing iteration-level scheduling mechanisms like Orca (\autoref{sec:eval:iteration-level})?
    \item What is the impact of our techniques on GPU bubbles and the throughput of pipeline-parallel models (\autoref{sec:eval:pp})?
    \item What are the overheads of \chunking (\autoref{sec:eval:ablation})?
\end{enumerate}

\input{5.1-eval-single-gpu}

\input{5.2-eval-multi-gpu}
\input{5.3-eval-ablation}

%% file: tables/tbl-eval-setup.tex
\begin{table}[t]
\centering
\setlength{\tabcolsep}{2pt}
\begin{tabular}{ccccc} \hline
Model & GPU & Num & Per-GPU & Mode \\
& & GPUs &  Mem(GB) & \\\toprule
LLaMA-13B & A6000 & 1 & 48 & Deployment \\
LLaMA-33B & A100 & 1 & 80 & Deployment \\
GPT-3 & A100 & 64 & 80 & Simulation \\
\bottomrule
\end{tabular}
\caption{Models, GPUs, and mode of evaluation.}
\label{tbl-eval-setup}
\end{table}

%% file: 5.1-eval-single-gpu.tex
\subsection{Evaluation on a Single GPU}
\label{sec:eval:end-to-end}

In this section, we measure the decode speedup and the end-to-end throughput of \sysname, on a single GPU, against that of the baseline which executes the prefill and decode stages separately via prefill-only and decode-only batches. Further, we examine the effects of varying $P:D$ ratio (ratio of prefill to decode tokens), sequence lengths (total tokens per request --- $P+D$), and batch sizes on the overall throughput.
\input{fig-tex/fig-eval-decode-speedup}

\subsubsection{Decode speedup}
\label{sec-eval-decodes}
We first show the impact of our techniques on decode phase throughput that we calculate based on the average time spent on decoding one token. For the baseline system, we compute the average decode time per token by dividing the time to process one decode iteration by the batch size. In \sysname, where decodes are piggybacked, for a batch with $p+d$ tokens, where $p$ denotes the prefill chunk size and $d$ denotes the decode batch size, we find the difference in runtime between the {\em decode-maximal} batch and a prefill-only batch of prefill size $p$, and attribute the difference in time as the marginal decode time for a batch of $d$ requests. This marginal decode time is then used to compute the decode time per token.

~\autoref{fig:eval:decodeefficiency} plots the results for a chunk size of 256 for LlaMa-13B on A6000 GPU, as we vary the batch size, up to the respective maximum value that fits, for three different prefill sequence lengths. We observe that \chunking improves decode efficiency by up to an order of magnitude over baseline. Decode throughput of \sysname is higher due to \hbatch that computes decode tokens with matrix-multiplications, allowing reuse of the model weights --- for both prefills and decodes --- once they are fetched from the  GPU's global memory.

We observe that our decode speedup reduces as we increase the batch size or sequence length. This behavior is expected for the following reasons: (1) decodes in the baseline system become more efficient as the batch size increases, and (2) the cost of attention increases quadratically with the sequence length: since all our improvements come from optimizing the linear operations, a higher attention cost reduces our scope for improvement. However, our decode throughput improvement is still significant in all cases ($2.8\times-10\times$).

\input{tables/tbl-eval-peak-perf}

\input{fig-tex/fig-eval-p-d}

\input{fig-tex/fig-eval-all-configs}

\subsubsection{Peak throughput gains with \sysname}
\autoref{table:eval:peak:throughput:combined} shows the peak throughput gain that \sysname achieves over the baseline.
To demonstrate the generality of our techniques, we evaluate \sysname on two model-GPU combinations: (1) LLaMA-13B on an A6000 GPU and (2) LLaMA-33B on an A100 GPU. Further, we investigate the peak throughput gain with varying sequences of length 1K, 2K and 3K. \autoref{table:eval:peak:throughput:combined} shows the batch sizes and P:D ratios where we achieve the maximum speedup.

In the best case, our techniques improve the end-to-end throughput by as much as $1.33\times$ for LLaMA-13B and up to $1.25\times$ for LLaMA-33B. We observe that the speed up is relatively higher on the A6000 GPU as compared to the A100 GPU. This is due to the higher FLOPs/MemBandwidth of the A100 GPU compared to the A6000 GPU ($\approx156$ vs. $\approx53$, ignoring GPU caches). Therefore, we require a higher chunk size on the A100 GPU (or a model with a higher embedding size) to avoid losing the prefill efficiency. However, \sysname still consistently outperforms the baseline by $1.14\times$-$1.25\times$ on the A100 GPU. These results show that piggybacking decode tokens with prefill chunks is useful across a wide range of models and hardware.  We note that although we improve decode efficiency by up to an order of magnitude, the end-to-end speedups and in turn monetary savings in inference cost are in the order of 25\%. This is because our technique only improves decodes and not prefills.

\subsubsection{Effect of varying $P:D$ ratio}
\label{sec-eval-e2e}
In this subsection, using various sequence lengths and chunk sizes, we investigate the effect of varying $P:D$ ratios on the end-to-end inference throughput to cover a wide range of application scenarios. $P:D$ ratio is an important parameter for these experiments: a lower $P:D$ ratio indicates that a request constitutes more decode tokens compared to other requests with a higher $P:D$ ratio. Although a lower $P:D$ ratio implies that decodes will constitute a larger fraction of the inference cost and thus \sysname will have more surface area of attack, however, it also means there will be fewer prefill chunks for piggybacking decodes. This trade-off results in a behavior where the improvement from \sysname peaks at a particular $P:D$ ratio and then tapers off on either side. We discuss this in more detail below.

\autoref{fig:eval:p:d} plots the results of our experiments. We find that the peak efficiency of our techniques occurs at different $P:D$ ratios for different prefill chunk size and batch size scenarios. If $C$ is the chunk size and $B$ is the batch size, then we can show that this peak will occur when the decodes perfectly piggyback with the prefill chunks. This occurs when the number of prefill chunks ($=P/C$) is the same as the required number of decode iterations (=$D/(B - 1)$), i.e., when $P:D = C/(B - 1)$. For example, using a chunk size of 256 at batch size of 18, \sysname achieves the peak throughput improvement of $1.27\times$ at $P:D=14$ ($\approx C/(B - 1) = 256 / 17$) for sequence length of 1K as shown in ~\autoref{fig:eval:pd:1K}.
Using the chunk size of 512 for sequence length=1K at batch size of 18 also provides significant gains of up to $1.23\times$ at $P:D=28$ ($\approx C/(B - 1) = 512 / 17$) whereas the gains are much lower with a chunk size of 128. While smaller chunks provide more opportunity to overlap decodes, splitting prefills into very small chunks leads to lower arithmetic intensity i.e. less efficient matmuls and higher overheads (due to multiple reads of KV cache), resulting in reduced end-to-end performance. Thus we obtain a much higher throughput with chunk size of 256/512 compared to the smaller chunk size of 128. Note that the peak gains occur at a higher value of $P:D$ ratio when using a larger chunk size.

We achieve peak performance when inference is not entirely dominated by either prefills or decodes (in other words, when the $P:D$ ratio is balanced). Such a state allows us to overlap prefills and decodes efficiently for longer. Otherwise, \sysname either runs out of prefill tokens (if $P:D$ is low) or decode tokens (if $P:D$ is high). In these cases, \sysname can switch to a different chunk size, or operate similar to the standard baseline processing prefill-only or decode-only batches. However, note that despite this variation, our improvements are still around 10\% over a large range of $P:D$ ratios.

\subsubsection{Effect of varying the batch and chunk sizes}
In this section, we dive deeper to investigate the performance of \sysname by varying the batch sizes and chunk sizes for each sequence length. In all these experiments, we focus on execution scenarios where the $P:D$ ratio is balanced i.e., when $P:D = C/(B - 1)$ and all decode tokens are perfectly piggybacked with prefills. This allows us to measure the peak performance of our system.

~\autoref{fig:all_configs} shows the results for these experiments. For each configuration of sequence length and chunk size, we show the effect of varying batch sizes. Further, for each run, we also show the runtime across different operations i.e., {\em preproj}, {\em attention}, {\em postproj}, and {\em ffn}.

Note that \hbatch batches the prefill and decode tokens in linear operations to improve compute utilization. Therefore, the linear operations see a significant runtime reduction of up to $1.6\times$ (see {\em ffn} runtime in the first row) compared to the baseline. However, note that the magnitude of improvement also depends on the $P:D$ ratio (in other words, it depends on what fraction of time is spent in decodes). For example, using a chunk size of 256 doubles the number of decodes that can be piggybacked compared to using 512 as the chunk size. Therefore, in the optimal configurations ($P:D = C/(B - 1)$), for chunk size of 256, decodes constitute a higher fraction of total runtime, compared to the optimal configuration when chunk size is 512. Therefore, our throughput gains are higher when using chunk size of 256.

We also observe that different linear operations see different speedups using our technique. Linear computation in the {\em ffn} module sees the highest runtime reduction of $1.3\times$-$1.6\times$. In contrast, the runtime reduction for {\em preproj} and {\em postproj} is $1.05\times$-$1.38\times$.
For small batch sizes, we find that most of the throughput improvement is due to the higher efficiency of {\em ffn} computation in \hbatch.

\input{fig-tex/fig-eval-iter-level}

\subsection{Comparison to Iteration-level Scheduling}
\label{sec:eval:iteration-level}

In our evaluation thus far, we have considered a baseline system that processes prefill-only or decode-only batches at a time. This is how popular frameworks like FasterTransformer deploy transformer models. In contrast, Orca's iteration-level scheduling~\cite{orca} can add (or remove) a request to (or from) a running batch at the granularity of individual iterations.

Iteration-level scheduling affects GPU utilization as well: when requests arrive or depart at different times, some prefills (of newly arriving requests) automatically overlap with the decodes (of already running requests). Therefore, we expect that iteration-level scheduling would do better than the baseline --- at least in some cases. However, we emphasize that the overlap between prefills and decodes is more of a side-effect in iteration-level scheduling and its behavior can vary significantly depending on the size and arrival or departure time of requests. Even more importantly, current approaches to iteration-level scheduling submit the entire input sequence of a request in a single prefill phase. This significantly limits the opportunity of piggybacking decode tokens with prefills.

To understand the effect on overall throughput, we evaluate the state-of-the-art iteration-level scheduler, Orca \cite{orca}, in two scenarios: its best-case and worst-case. In the best case, Orca scheduling overlaps the {\em full} prefill of {\em one} new request with the ongoing decodes. In the worst-case, all the requests begin and end at the same time. In the latter case, Orca scheduling behaves similar to our earlier baseline where there is no overlap between the computation of prefill and decode tokens. Note that in the average case of Orca, there could be more than one {\em full} prefill (corresponding to multiple requests) overlapping with some decodes -- this would further limit Orca's ability to piggyback decodes tokens with prefills.

~\autoref{fig:eval:orca} shows our results for these experiments. First (\autoref{fig:eval:orca}a), we show results for the optimal choice of $P:D = C/(B-1)$, where $C=256$ and $B$ is the maximum batch size that fits for the sequence length. As expected, worst-case Orca scheduling performs similar to the baseline. We find that, for a small sequence length of 1K, the best-case Orca scheduling achieves 1.11\myx higher throughput. This is due to the incidental overlapping of the prefill and decode requests in the best-case schedule. However, as sequence length increases, the performance of best-case Orca scheduling drops close to the baseline. This is an artifact of our choice of $P:D = C/(B - 1)$. As we increase sequence length, the batch size $B$ reduces, resulting in a higher optimal $P:D$. Since Orca submits the entire input sequence as a single prefill request, a higher $P:D$ means that it soon runs out of the prefill tokens, at which point it processes the remaining decode tokens similar to the baseline, making even the best-case version inefficient. \sysname consistently outperforms with overall throughput gains of 1.27\myx, 1.25\myx and 1.23\myx for the three sequence lengths.

Another aspect to consider in iteration-level scheduling is the effect of variable sequence lengths on request latencies. Since the prefill time increases with the length of the input sequence, adding a longer prefill sequence in a running batch can delay the ongoing decodes, which in turn increases the latency of these ongoing requests in Orca scheduling. \sysname avoids this due to the use of smaller chunk prefills.

Next, we evaluate the throughput gains at different $P:D$ ratios for different chunk sizes in ~\autoref{fig:eval:orca}b. We consider only sequence length of 1K for this experiment as the best-case Orca baseline achieves maximum performance in this regime. Note that best-case Orca scheduling can be considered a special case of \sysname, where the chunk size, $C$, is set to the maximum sequence length. As can be seen, the optimal $P:D$ shifts to the right as chunk-size increases. \sysname with chunk size of 256 performs the best in lower $P:D$ regimes, reaching a peak throughput gain of 1.27\myx compared to baseline. \sysname with chunk size of 512 consistently outperforms Orca best-case and performs overall best in the higher $P:D$ regime, reaching a peak throughput gain of 1.23\myx. In comparison, Orca best-case has much flatter gains and reaches a peak throughput gain of 1.11\myx at a much higher $P:D$.

\input{fig-tex/fig-eval-pp}

%% file: fig-tex/fig-eval-decode-speedup.tex
\begin{figure}
    \centering
    \includegraphics[trim={70 0 10 50}, clip, scale=0.26]{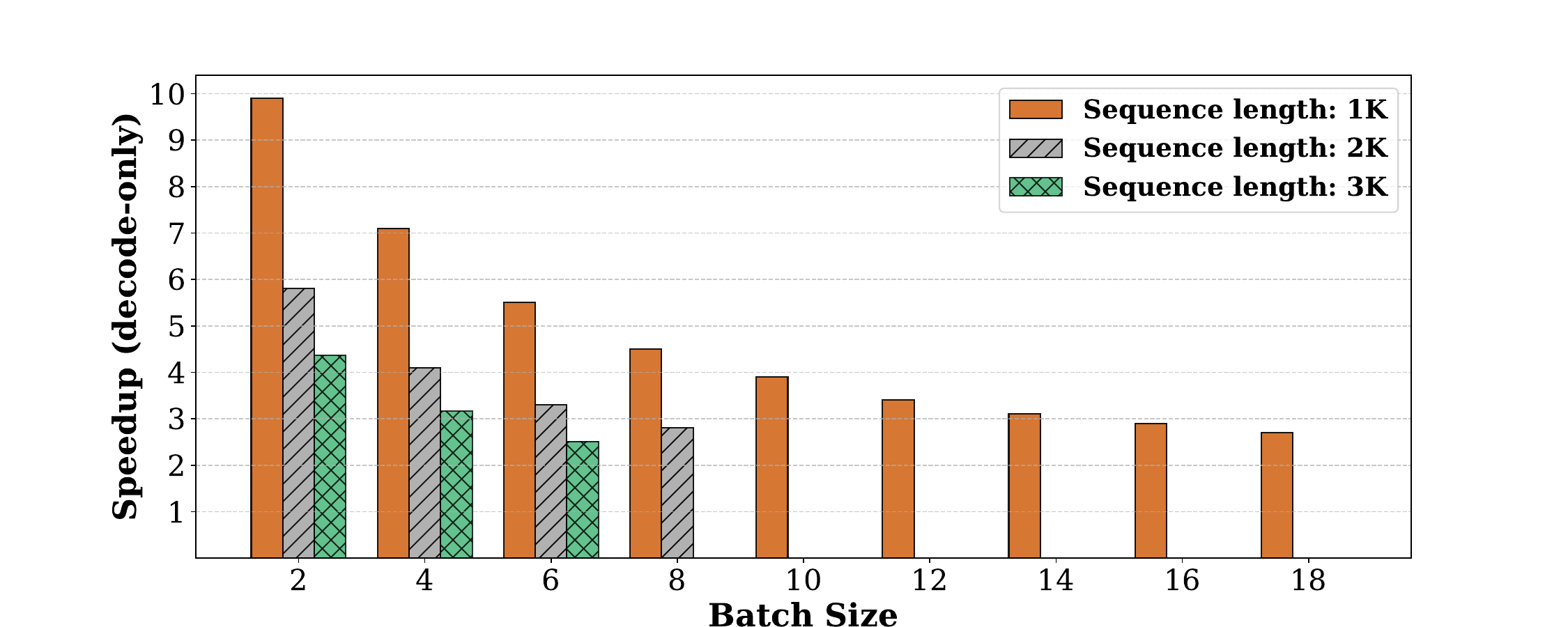}
    \caption{Decode-only speedup with \sysname on an A6000 GPU with LLaMA-13B (chunk size = 256).}
    \label{fig:eval:decodeefficiency}
\end{figure}

%% file: tables/tbl-eval-peak-perf.tex
\begin{table}
\center
\scalebox{0.8}{
\begin{tabular}{c|c|c|c|c|c}
{\bf Model} & {\bf Sequence} & {\bf Batch} & {\bf P:D} & {\bf Decode} & {\bf Throughput} \\ 
{\bf (GPU)} & {\bf Length}  & {\bf Size} & {\bf Ratio} & {\bf Speedup} & {\bf Gain} \\ \hline
LLaMA-13B & 1K & 6 & 50:1  & $5.45\times$ & $1.33\times$ \\ 
(A6000) & 2K & 6 & 50:1 & $3.26\times$ & $1.26\times$ \\ 
 & 3K & 6 & 50:1 & $2.51\times$ & $1.22\times$ \\ \hline
LLaMA-33B & 1K & 10 & 28:1 & $3.83\times$ & $1.25\times$ \\
(A100) & 2K & 5 & 63:1 & $4.25\times$ & $1.22\times$ \\ 
& 3K & 3 & 127:1 & $3.51\times$ & $1.14\times$ \\  \hline
\end{tabular}}
\caption{Peak throughput gains with \sysname for different sequence lengths with two different model-GPU combinations (chunk size = 256).}
\label{table:eval:peak:throughput:combined}
\end{table}

%% file: fig-tex/fig-eval-p-d.tex
\begin{figure*}[t!]%
    \begin{subfigure}[b]{0.33\textwidth}
    \centering
        \includegraphics[scale=0.20]{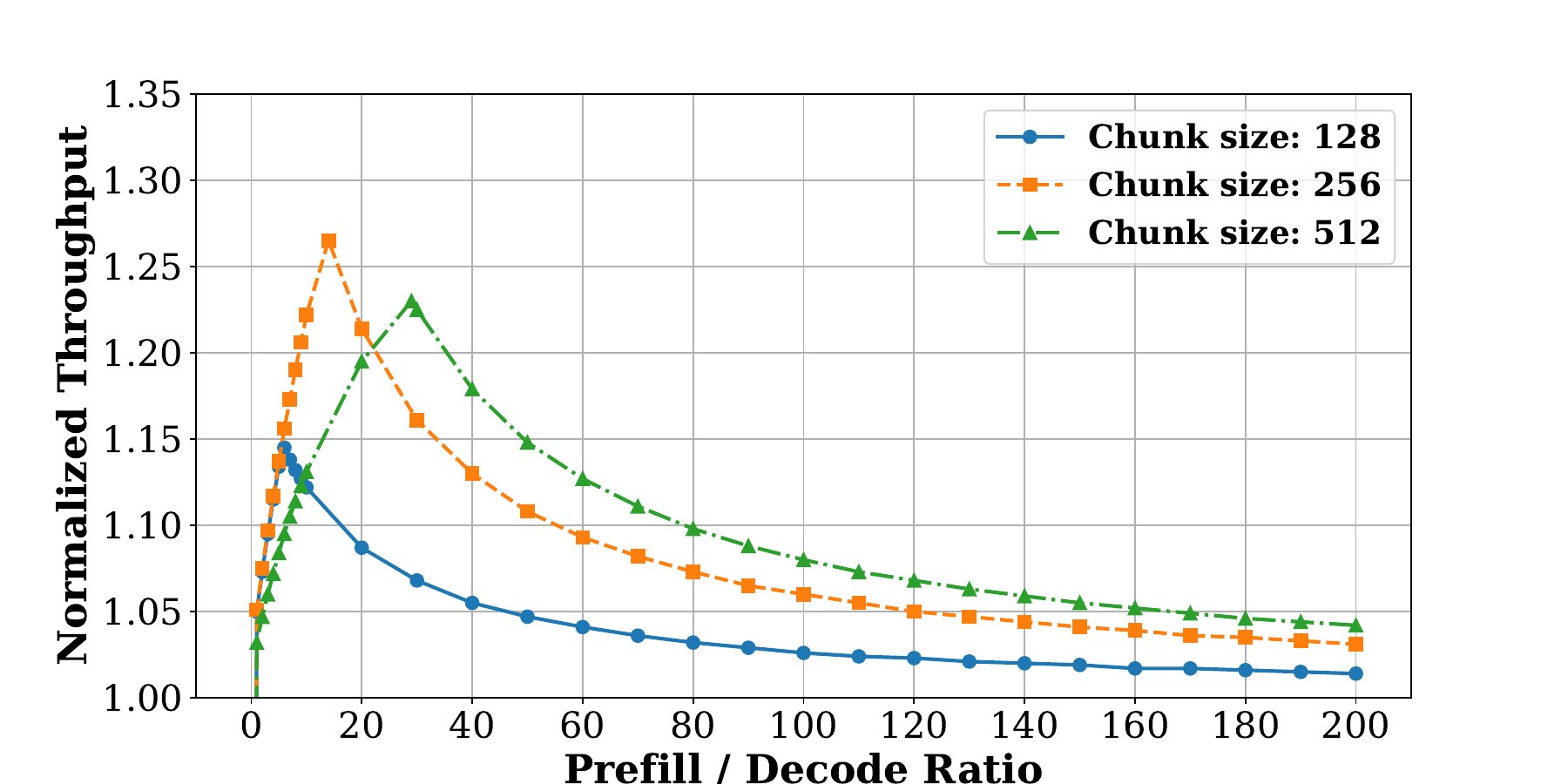}
        \caption{Sequence length = 1K, batch size = 18.}
        \label{fig:eval:pd:1K}
    \end{subfigure}
    \begin{subfigure}[b]{0.33\textwidth}
        \includegraphics[scale=0.20]{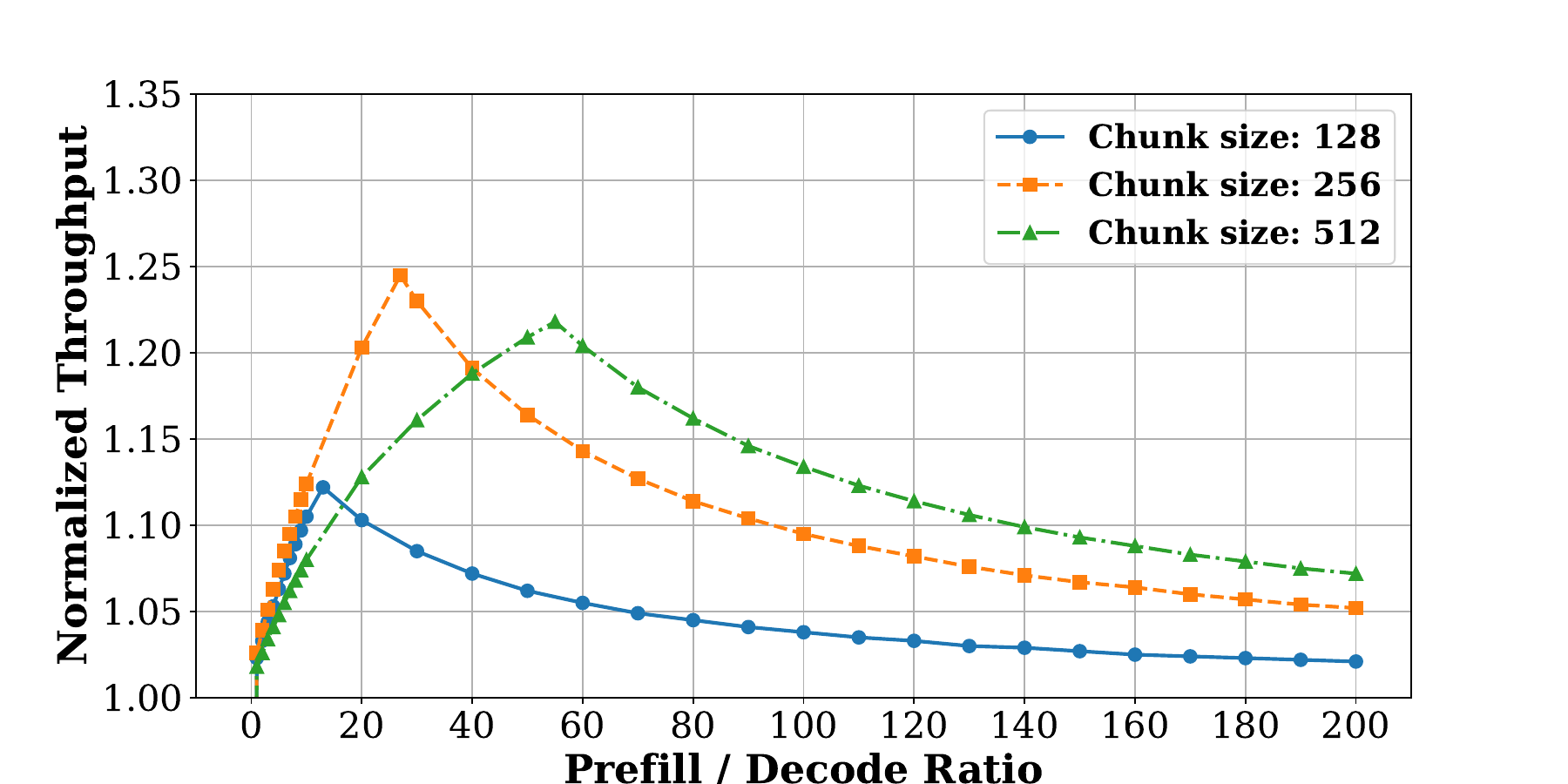}
        \caption{Sequence length = 2K, batch size = 10.}
        \label{fig:eval:pd:2K}
    \end{subfigure}
        \begin{subfigure}[b]{0.33\textwidth}
        \includegraphics[scale=0.20]{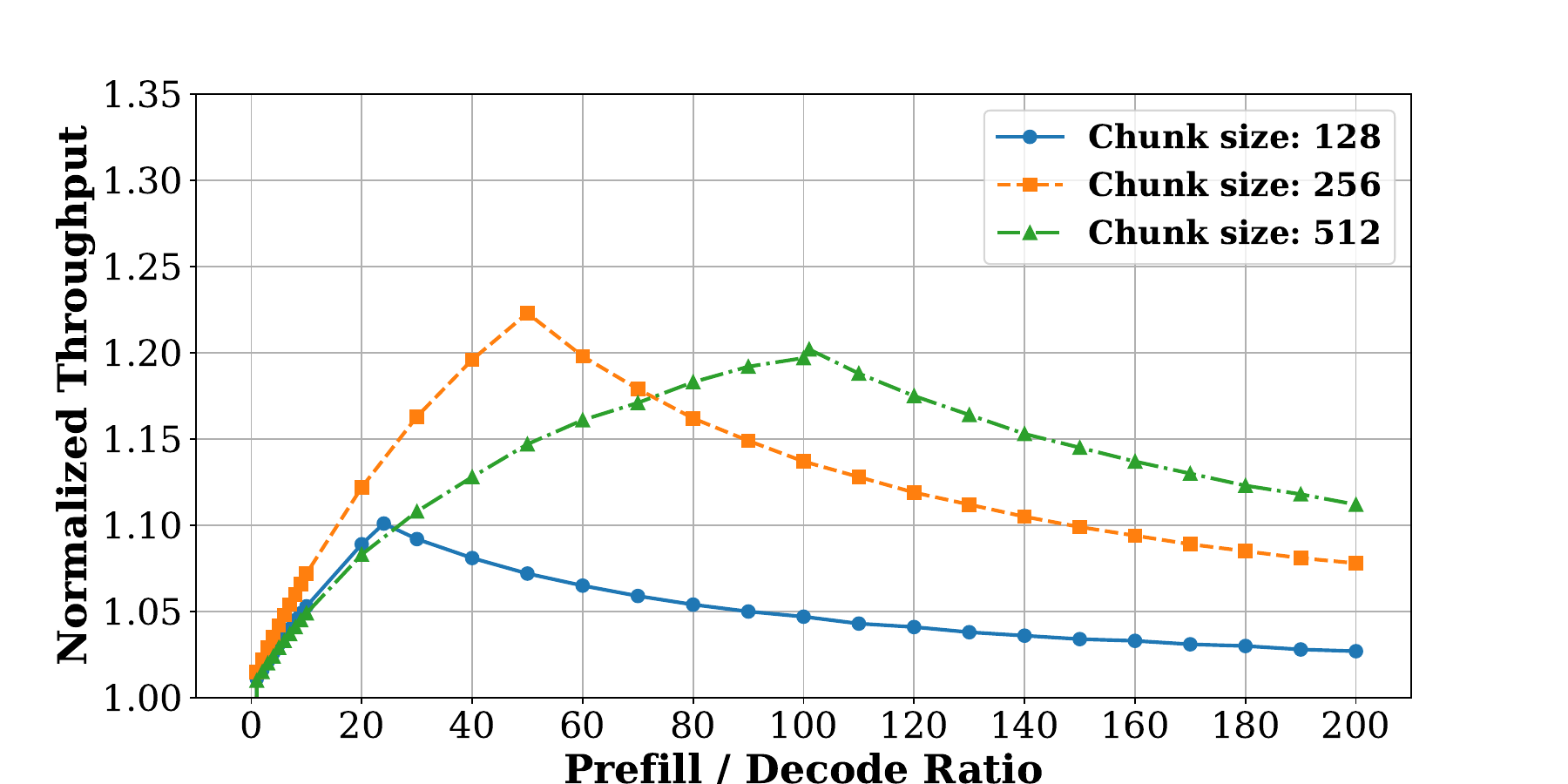}
        \caption{Sequence length = 3K, batch size = 6.}
        \label{fig:eval:pd:3K}
    \end{subfigure}
    \caption{Normalized throughput (tokens/ms) for LLaMa 13B on A6000 GPU with different sequence lengths, P:D ratios, and chunk sizes.}
    \label{fig:eval:p:d}
\end{figure*}

%% file: fig-tex/fig-eval-all-configs.tex
\begin{figure*}[!t]
    \centering
    \includegraphics[trim={0 50 0 0}, clip, scale=0.38]{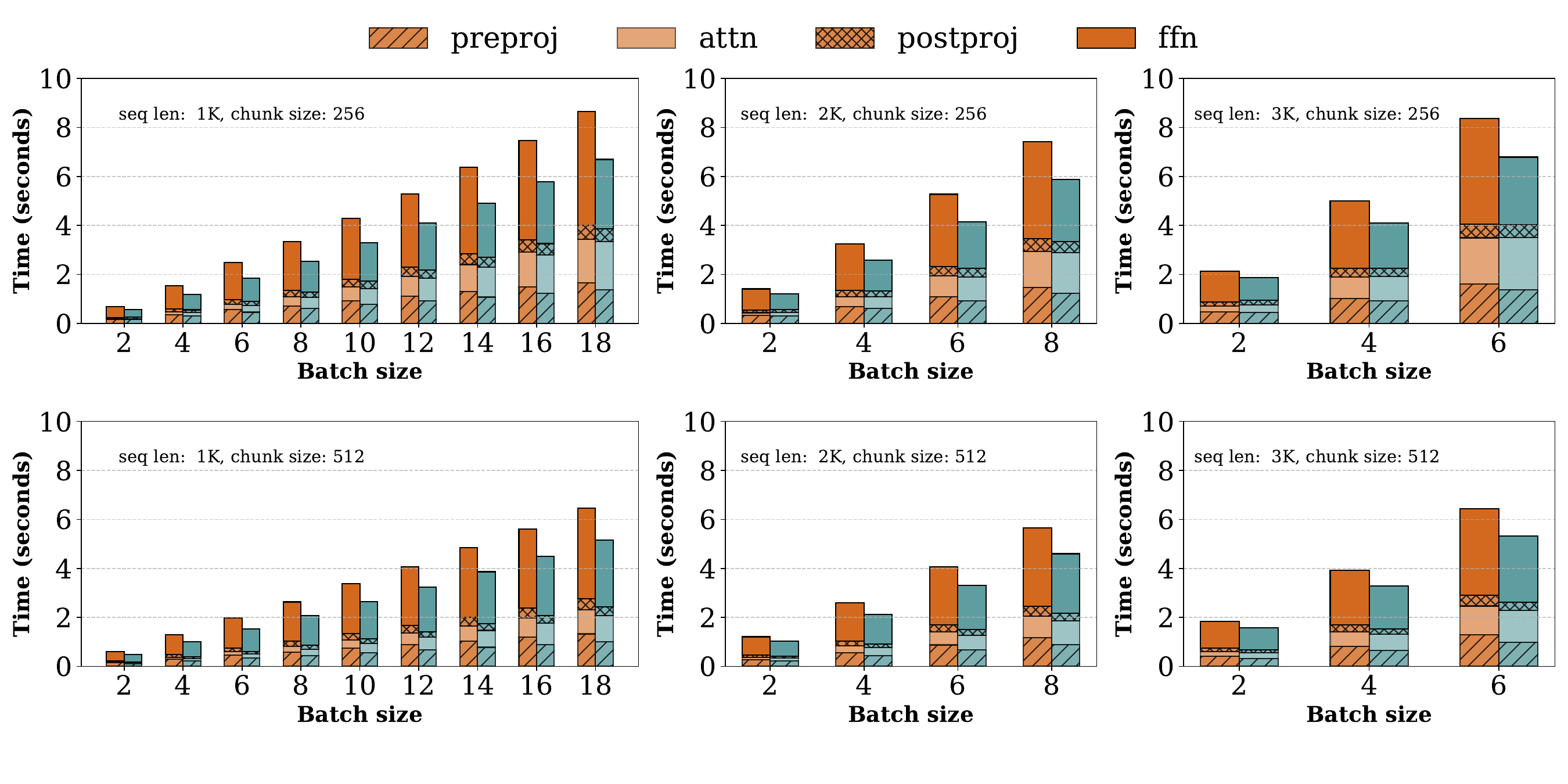}

    \caption{Breakdown of total time spent on different operations for LLaMa 13B on A6000 GPU with varying sequence lengths and batch sizes, using prefill chunk sizes of 256 (top half) and 512 (bottom half). Orange and blue bars represent baseline and \sysname, respectively.}
    \label{fig:all_configs}
\end{figure*}

%% file: fig-tex/fig-eval-iter-level.tex
\begin{figure}[!t]
    \centering
        \begin{subfigure}[b]{0.42\textwidth}
        \includegraphics[width=\textwidth]{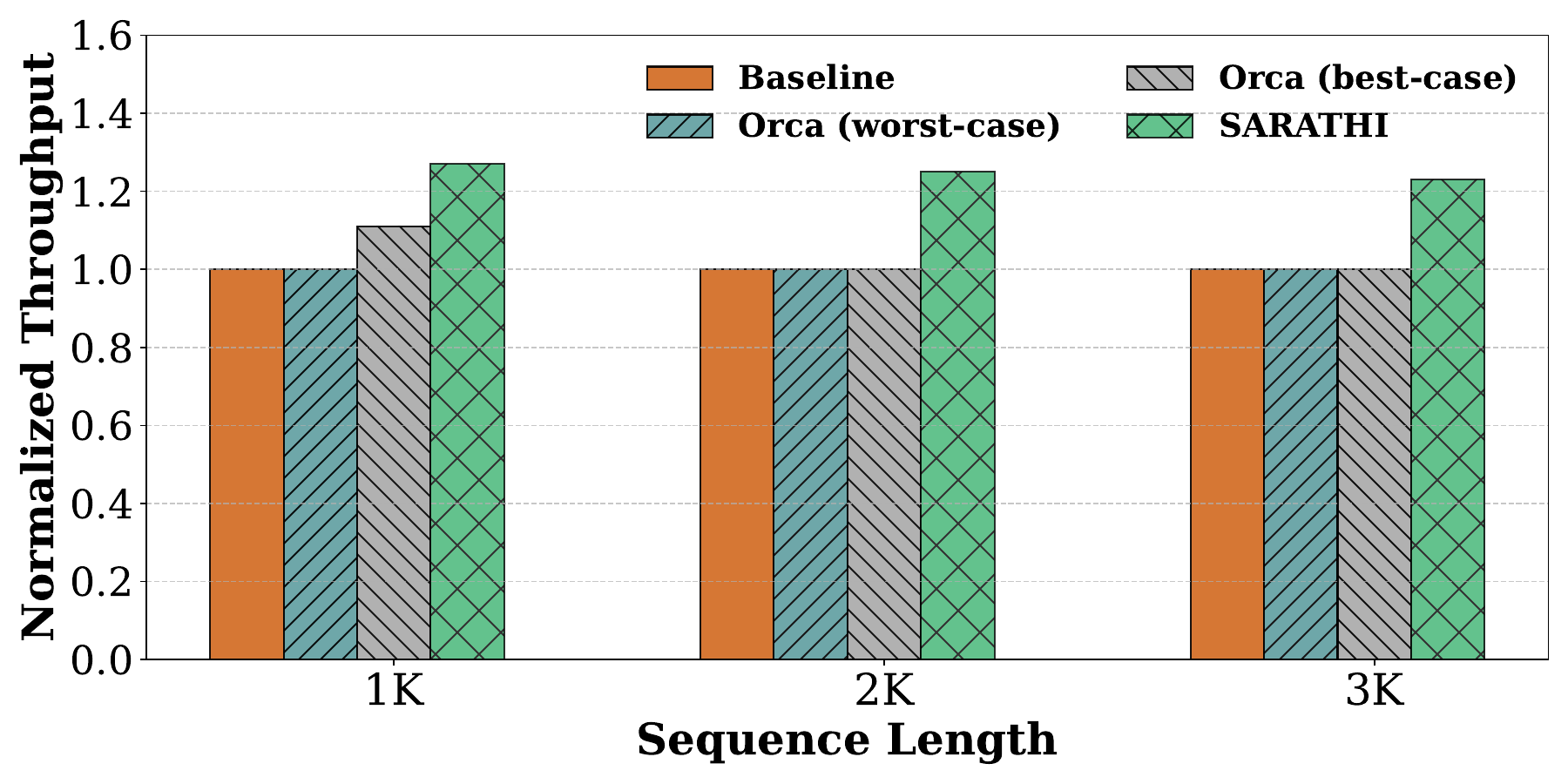}
        \caption{Varying sequence lengths (chunk size=256 for \sysname). We choose the maximum batch size which fits for the sequence length (18, 10 and 6 for 1K, 2K and 3K sequence lengths, respectively)}
        \label{fig:eval:orca:sequencelengths}
    \end{subfigure}
    \begin{subfigure}[b]{0.48\textwidth}
        \includegraphics[width=\textwidth]{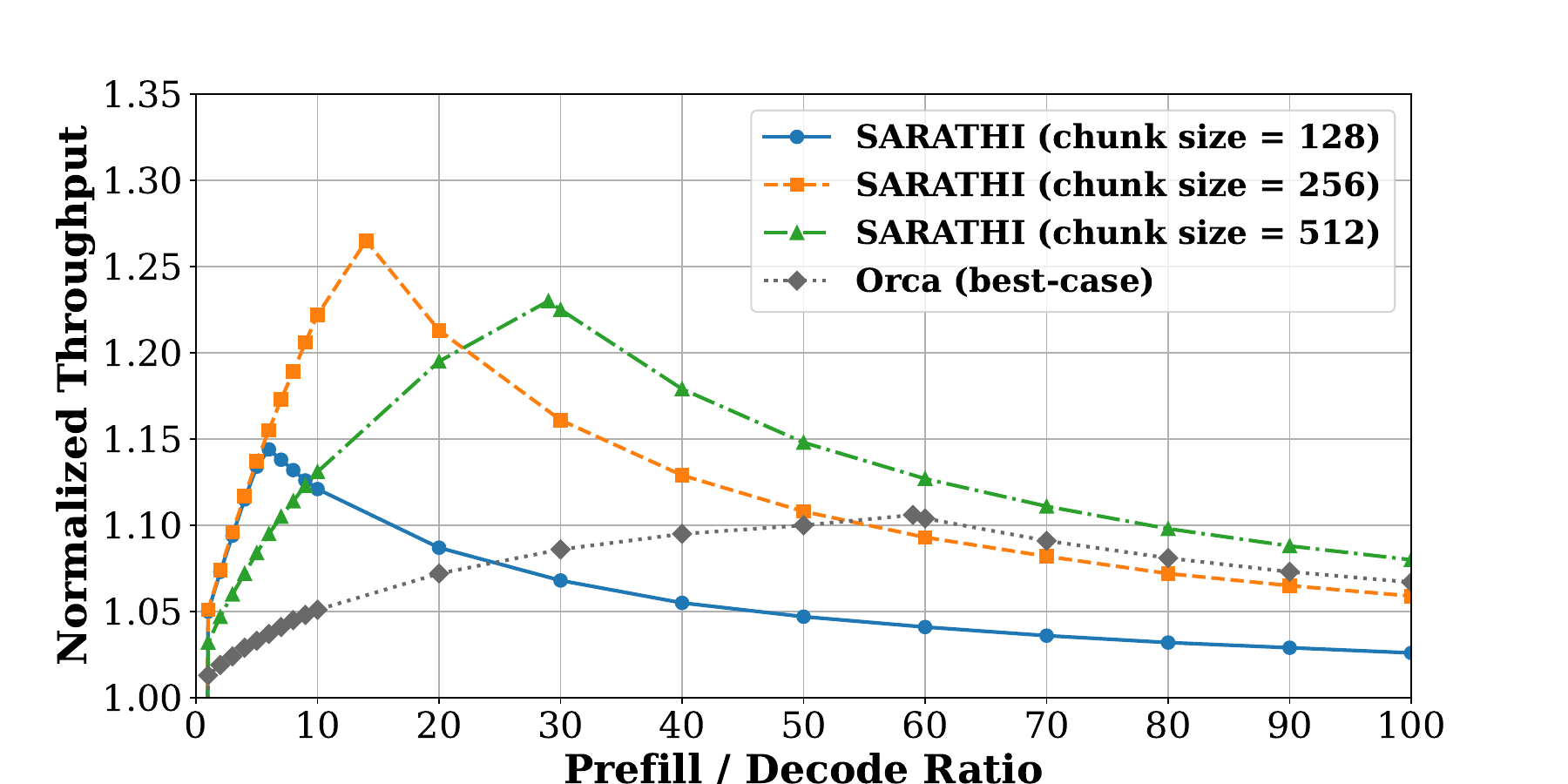}
        \caption{Varying P:D ratio (sequence length=1K, batch size=18).}
        \label{fig:eval:orca:pd:curve}
    \end{subfigure}
    \caption{Comparison with iteration-level scheduler Orca for LLaMa 13B on A6000 GPU.}
    \label{fig:eval:orca}
\end{figure}

%% file: fig-tex/fig-eval-pp.tex
\begin{figure}[tbp]
    \centering
    \begin{subfigure}[b]{0.35\textwidth}
        \includegraphics[width=\textwidth]{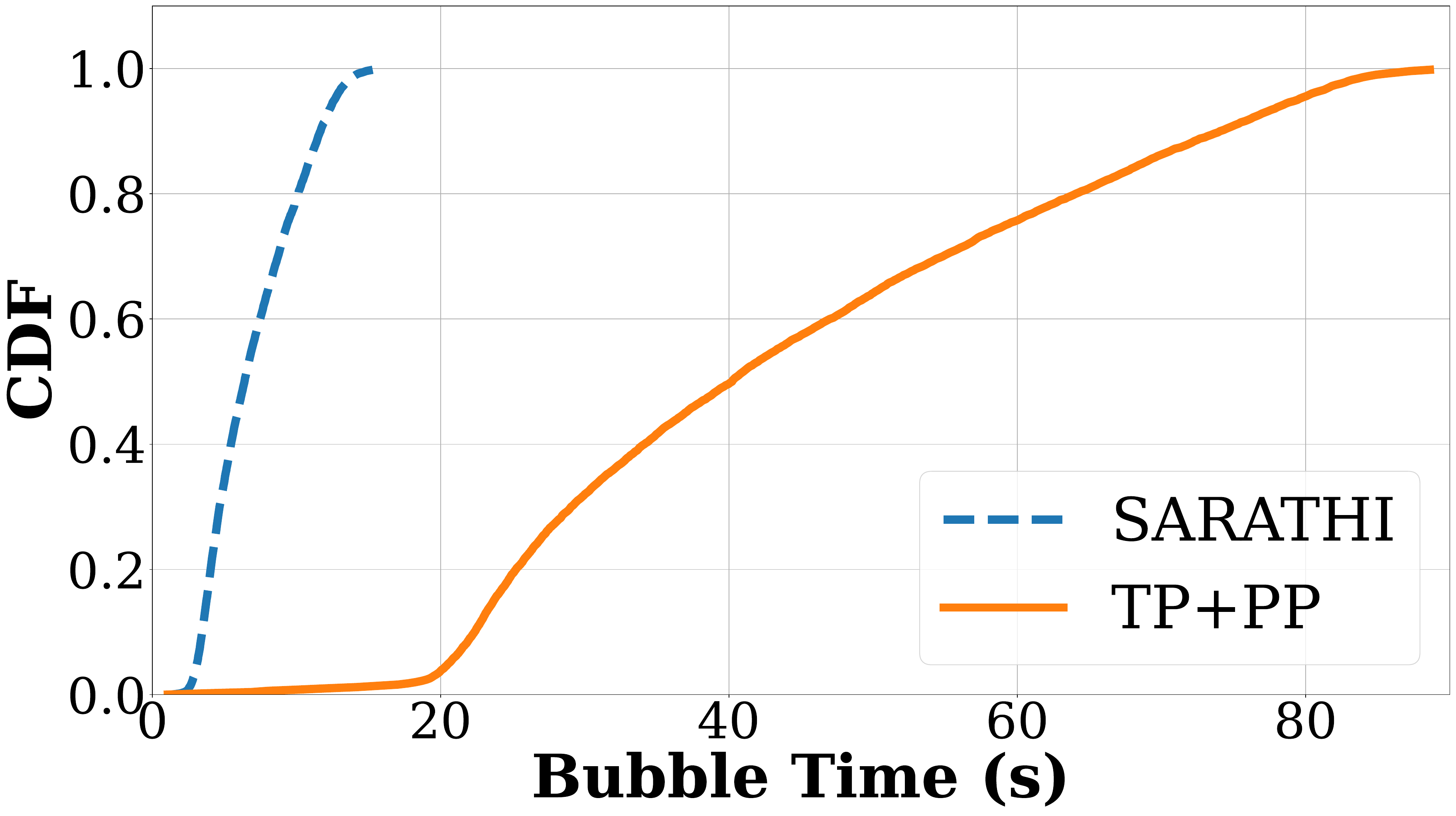}
        \caption{Comparison of bubble time}
        \label{fig-eval-pp-bubble}
    \end{subfigure}
    \\
    \begin{subfigure}[b]{0.37\textwidth}
        \includegraphics[width=\textwidth]{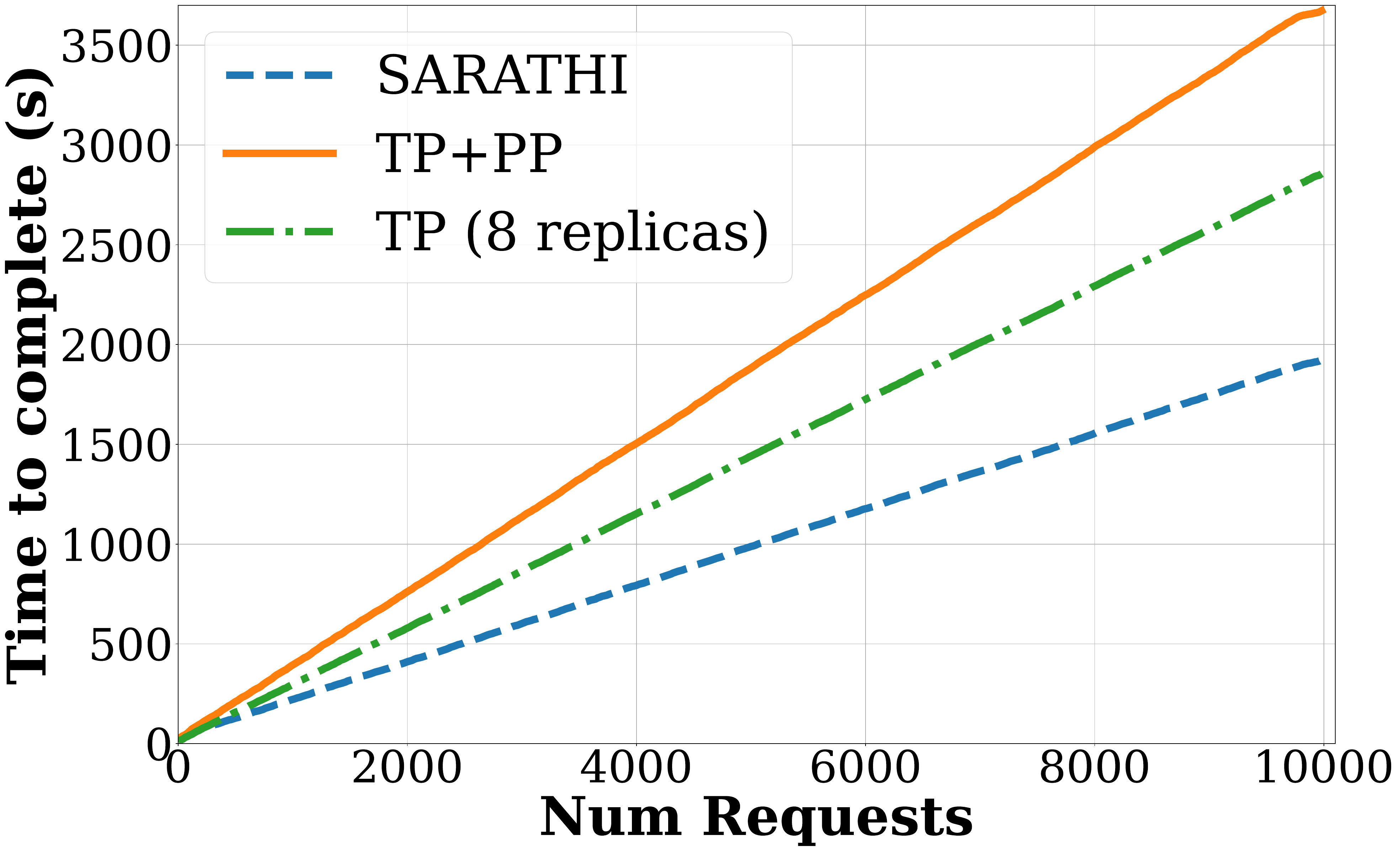}
        \caption{End-to-end request completion time}
        \label{fig-eval-pp-makespan}
    \end{subfigure}
    \caption{Impact of \sysname on pipeline bubbles (top) and request completion times (bottom) for GPT-3 deployed on DGX A100(s) in simulation.}
    \label{fig-eval-pp}
\end{figure}

%% file: 5.2-eval-multi-gpu.tex
\subsection{Pipeline Parallelism with \sysname}
\label{sec:eval:pp}

Next, we evaluate how \sysname reduces pipeline bubbles in a multi-GPU pipeline-parallel setup and subsequently impacts the overall runtime of inference jobs. For this experiment, we report evaluations in a carefully simulated environment. 

We first profile the runtime for each operation in ~\autoref{table:tensor:shapes} in the prefill and decode phase for various batch sizes and sequence lengths for the GPT-3 model~\cite{gpt3-brown2020language}. We further profile the network communication cost to faithfully simulate tensor-parallel and pipeline-parallel executions. Finally, we build a regression model to extrapolate and predict these values for missing data points that may be encountered during an online simulated inference serving system. We confirmed that the estimated runtimes by the simulator are within 5\% of the empirical values on an 8-GPU, 80GB A100 DGX box.

We report results for deployment over 64 A100 GPUs across eight servers connected with InfiniBand. We evaluate three scenarios; (1) 8-way tensor-parallel (TP) within a node with 8-way pipeline-parallel (PP) across nodes with the best-case Orca-style scheduling, (2) the same TP-PP setup as above with scheduling using \sysname's \chunking and \hbatch, (3) 8 parallel replicas, each with 8-way TP, serving simultaneously. For all scenarios, we use the maximum batch size that fits the GPU --- for TP+PP this was 27 and for TP only this was 11. The P:D ratio is fixed at 10 for this simulation with the minimum and maximum sequence length of the requests set to 1K and 4K respectively. %
Each request may have a different sequence length which is sampled from a Zipf distribution ($\theta = 0.4$), adhering to the maximum sequence length. The number of prefill and decode tokens is then calculated by satisfying the desired P:D ratio. For this experiment, we set the chunk size to be 256.

\input{fig-tex/fig-eval-chunked-prefills}

~\autoref{fig-eval-pp-bubble} plots the cdf of pipeline bubble time per request. We define this as the sum of bubble time for all the micro-batches across all iterations for a given request. \sysname reduces the median bubble time per request by 6.29\myx, by creating equal-compute units of work.

Next, we compare the overall request completion time for the different scenarios 
in ~\autoref{fig-eval-pp-makespan}. This graph plots the time to complete a given number of requests (our simulation considers a total of 10K requests). The TP-PP execution requires less memory for storing parameters compared to the TP-only setup, resulting in more room for the KV cache. Thus the TP-PP deployment supports 2.45\myx higher batch size compared to TP-only deployment, and yet, we observe that the TP-only execution is 1.28\myx faster than the baseline TP-PP with Orca scheduling, due to the large pipeline bubbles in the latter case. However, with \chunking and \hbatch, \sysname enabled PP execution is accelerated by 1.91\myx compared to the baseline TP-PP, and by 1.48\myx compared to the TP-only execution. 
Thus, \sysname makes pipeline parallel execution an attractive option for LLM inference by significantly minimizing pipeline bubbles.

%% file: fig-tex/fig-eval-chunked-prefills.tex
\begin{figure*}[t!]%
    \begin{subfigure}[b]{0.33\textwidth}
    \centering
        \includegraphics[scale=0.20]{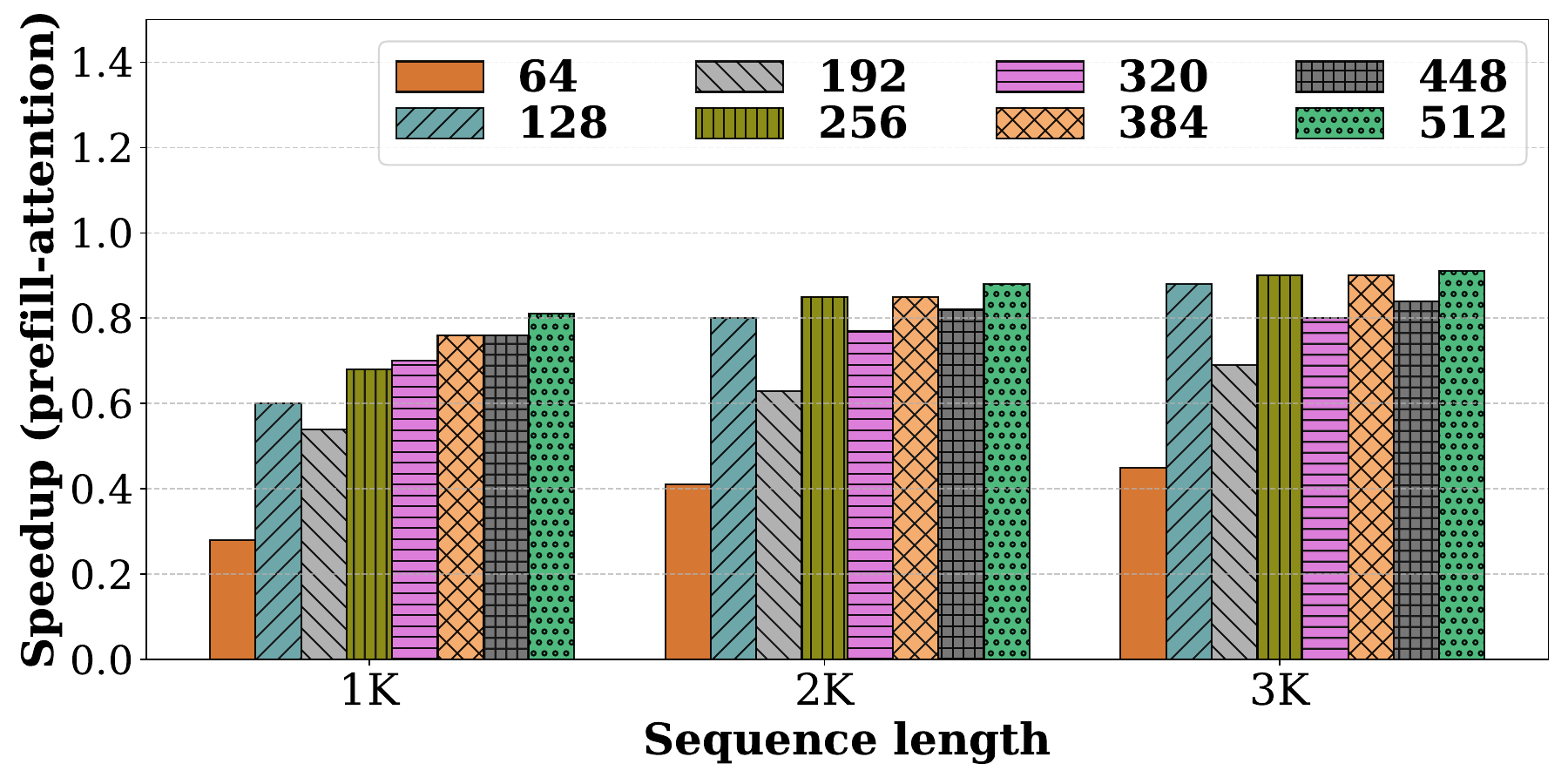}
        \caption{Self-attention (prefill-only).}
        \label{fig:eval_chunk_prefills_attention}
    \end{subfigure}
    \begin{subfigure}[b]{0.33\textwidth}
        \includegraphics[scale=0.20]{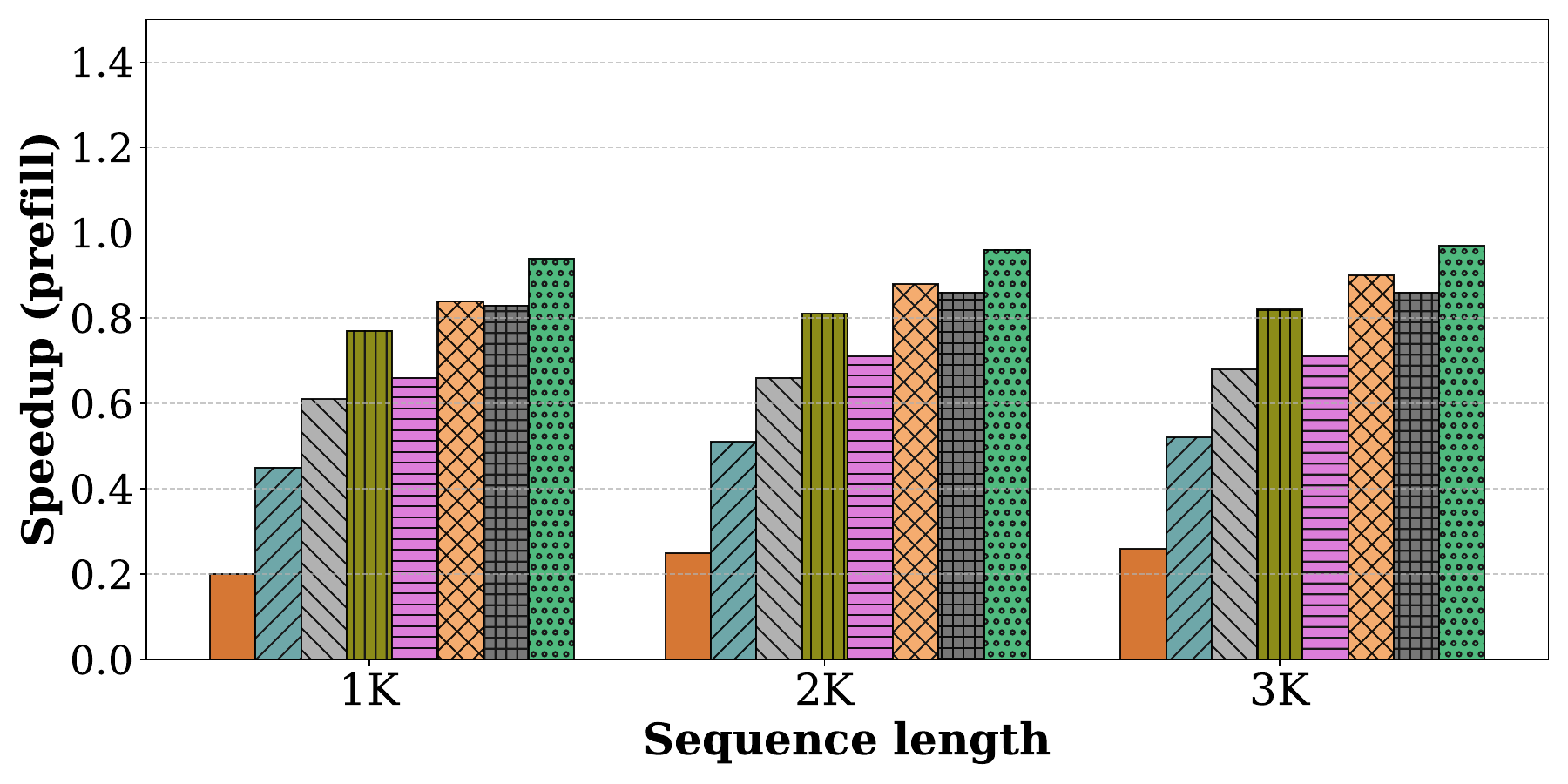}
        \caption{chunked-prefills vs. full prefill.}
        \label{fig:eval_chunk_prefills_only}
    \end{subfigure}
        \begin{subfigure}[b]{0.33\textwidth}
        \includegraphics[scale=0.20]{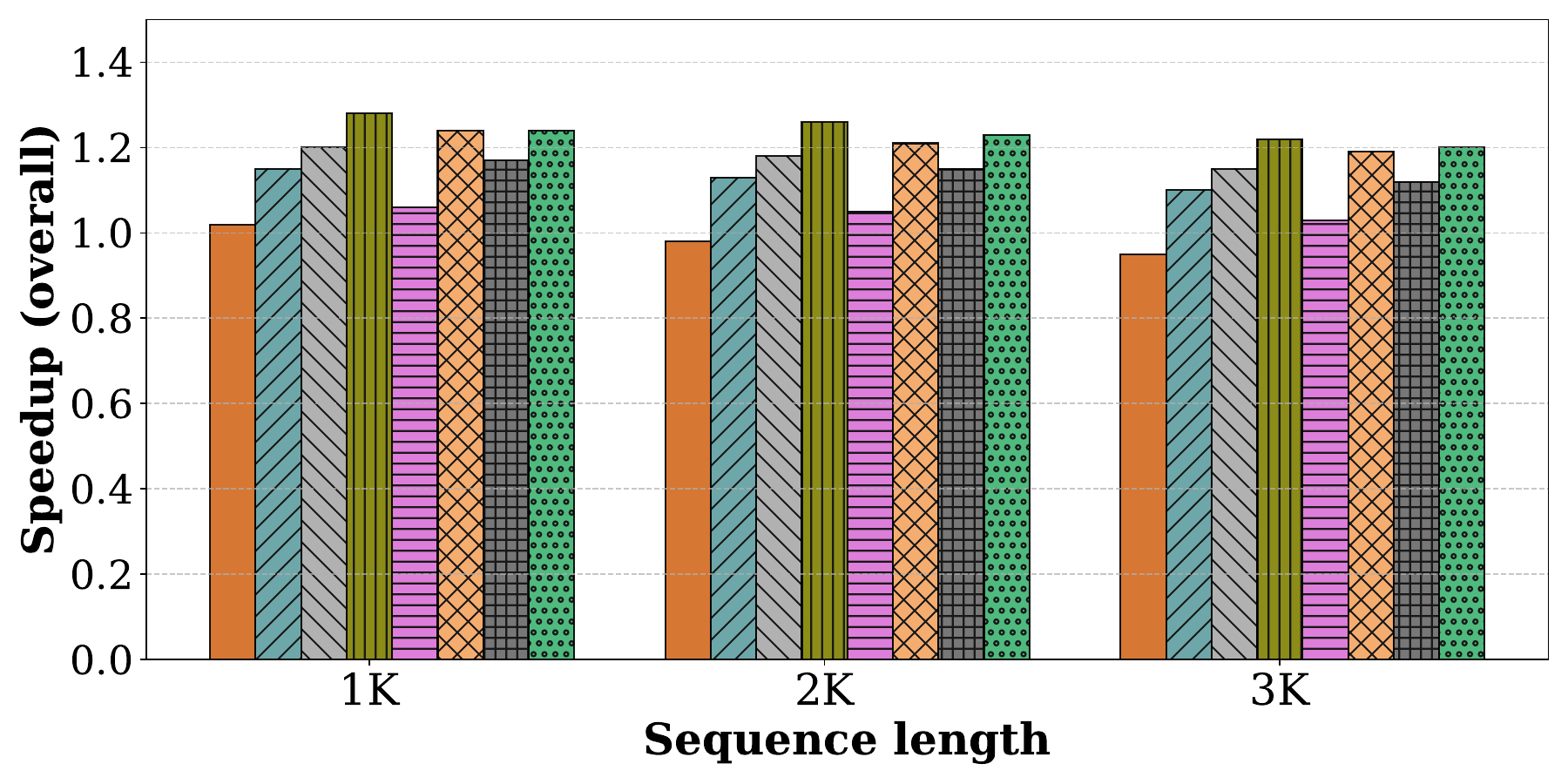}
        \caption{End-to-end speedup for the entire batch.}
        \label{fig:eval_chunk_prefills_overall}
    \end{subfigure}
    \caption{{\bf Ablation study:} Effect of varying the chunk size on different components of the system for LLaMa 13B on A6000 GPU.}
    \label{fig:eval_chunk_prefills}
\end{figure*}

%% file: 5.3-eval-ablation.tex
\subsection{Ablation Study of \Chunking}
\label{sec:eval:ablation}

In this subsection, we evaluate how splitting a full prefill computation into multiple smaller prefill chunks affects the efficiency of the prefill stage in \sysname. To quantify this, we measure the time to compute the prefill phase for various sequence lengths using the full sequence at once - this represents our baseline prefill performance. For each long sequence, we then compute the prefill with \chunking and compare its end-to-end runtime with the baseline. The difference between the two indicates the overhead of \chunking.

Prefill chunking has two potential sources of overheads: (1) it uses smaller chunk sizes compared to the baseline which may lower the GPU utilization, and (2) it needs to load the KV cache of each chunk multiple times, depending on the number of chunks in a request. Therefore, to fully understand the overhead of prefill chunking, we investigate the following: (1) what is the impact of chunking on attention computation for a prefill-only batch, (2) what is the effect of chunking on the overall runtime of prefill-only batch, and (3) what is the end-to-end throughput when \chunking is used in tandem with \hbatch. We study these by varying the chunk size from 64 to 512 as shown in~\autoref{fig:eval_chunk_prefills}.

First, we observe that smaller chunk sizes can add significant overhead, for both attention and the overall prefill runtime. For example, the chunk size of 64 incurs $3\times$ overhead for attention (see~\autoref{fig:eval_chunk_prefills_attention}) and about $5\times$(see~\autoref{fig:eval_chunk_prefills_only}) in the overall prefill time. As one can expect, the overhead of \chunking is lower for large chunk sizes: this is a combined effect of higher GPU utilization and fewer KV cache reloads with larger chunks. Overall, we find that chunk sizes of 256 and 512 provide reasonable prefill efficiency, limiting the end-to-end prefill computation loss to within 20\% and 10\%, respectively. 

Second, \sysname can compensate for some loss in prefill efficiency by improving the decode throughput. For instance, we see from ~\autoref{fig:eval_chunk_prefills_overall} that a chunk size of 64 almost matches the performance of our baseline despite being $5\times$ slower in prefill whereas a chunk size of 128 yields up to $1.16\times$ higher throughput despite its prefill being more than $2\times$ slower than the baseline, mainly due to piggybacking more decodes.  The tile-quantization effect is also evident in \autoref{fig:eval_chunk_prefills} as \sysname achieves higher improvement in throughput when the chunk size is a multiple of 128; e.g., chunk size 256 shows better speedup than 320.

%% file: 6-discussion.tex
\section{Discussion} \label{sec:discussion}
In this paper, we have comprehensively demonstrated how \sysname improves the performance of LLM inference across several models and hardware configurations. However, there are multiple challenges that require further investigation.

First, we focus only on an efficient scheduling mechanism in \sysname to improve the throughput of LLM inference. However, real-world deployments need to optimize an inference serving infrastructure simultaneously along multiple dimensions e.g., latency, queuing delays, fairness, etc. Meeting these goals with \sysname requires revisiting scheduling policies. Second, although we show what is an appropriate chunk size for a given P:D ratio, we leave it to future work to explore how to pick an optimal chunk size as it depends on several factors like the hardware, model characteristics, sequence length, and the composition of prefill-decode tokens, especially in scenarios where the P:D ratio may not be known ahead of time. Third, we make a simplistic assumption in this paper that each request in a batch has the same number of prefill and decode tokens (except the simulation experiments) whereas, in the real world, the sequence lengths can vary significantly across different LLM inference requests. Finally, we focused on sequence lengths of up to 3K, and P:D ratio in the range of 1-200. We believe that these are representative of many real-world deployments. However, there has also been an increased interest in supporting very long sequences (e.g., 10s-100s of thousands~\cite{largecontextlength}). Such large sequence lengths may pose new challenges as the cost of attention grows quadratically with the number of tokens. We are actively investigating these challenges.

%% file: 7-related.tex
\section{Related Work}
In this section, we provide a brief summary of related work along two dimensions: systems optimizations and model innovations.

\subsection{\bf Systems Optimizations} 

\noindent{\bf Memory management:} In auto-regressive decoding, the number of tokens that need to be generated for a given request is not known apriori. Therefore, conventional systems pre-allocate memory for the KV cache based on a conservative estimation of the maximum number of tokens. Recently, vLLM showed that this approach is inefficient and proposed a framework --- motivated by the virtual memory abstraction --- that enables incremental memory allocation for KV caches~\cite{vLLM:github}. This helps improve the batch size, especially when the number of tokens varies significantly across different requests. FlexGen~\cite{flexgen} focuses on improving the throughput of offline LLM inference in resource-constrained scenarios e.g., running a large model on a single GPU. Toward this goal, FlexGen employs a judicious combination of memory offloading, quantization, and scheduling.

\noindent{\bf Optimizing (self-)attention: } In ~\cite{self:attn:on2:memory}, the authors propose an algorithm to reduce the memory requirement of self-attention from $O(n^2)$ to $O(1)$, with respect to the sequence length. FlashAttention~\cite{flashattention} proposed a tiling-based algorithm that speeds up attention computation by minimizing the number of bytes read/written between different levels of GPU memory. Follow-up work~\cite{flashattention2} on FlashAttention further improved it along parallelism and work partitioning~\cite{flashattention2}. In our experiments, we found the xformers memory efficient attention implementation~\cite{xformers} to be the most efficient.

\noindent{\bf Kernel-level optimizations:} FasterTransformer~\cite{fastertransformer} proposed optimized layers for the transformer's encoder and decoder blocks. These are based on low-level GPU optimizations such as kernel fusion.
We expect that such low-level optimizations would equally benefit \sysname as well.

\noindent{\bf Scheduling optimizations:} Orca proposed an iteration-level scheduling framework that avoids wasting compute due to token padding that was used earlier to batch together requests with different sequence lengths~\cite{orca}. Further, Orca reduces latency by returning the response as soon as a request's end-of-sequence token gets generated. FastServe proposed a preemptive scheduling framework to minimize the job completion times~\cite{fastserve}. Some other scheduling frameworks include Triton~\cite{triton} and Clipper~\cite{clipper} that separate the serving layer from the execution engine of the model. Our current work focuses on optimizing the execution layer and can be used with different scheduling policies proposed by such systems.

The optimizations proposed by several of the prior works can complement our optimizations e.g., more optimized attention implementations will enable scaling \sysname to longer sequence lengths and dynamic memory allocation will help in supporting larger batch sizes and so on.

\subsection{Model Innovations} A significant body of work around model innovations has attempted to address the shortcomings of  transformer-based language models or to take the next leap forward in model architectures, beyond transformers. For example, multi-query attention shares the same keys and values across all the attention heads to reduce the size of the KV cache~\cite{multiqueryattention}, allowing larger batch sizes. Several recent works have also shown that the model sizes can be compressed significantly using quantization~\cite{smoothquant,gptq,qlora,llmint8}. Mixture-of-expert models are aimed primarily at reducing the number of model parameters that get activated in an iteration~\cite{largescalemoe,moeatc23,moedeployment}. More recently, retentive networks have been proposed as a successor to transformers~\cite{retnet}. In this work, we focus on addressing the performance issues of the most popular transformer models from a GPU's perspective.  Model innovations are orthogonal to our work.

%% file: 8-conc.tex
\section{Conclusion}
\label{sec-conclusion}

In this paper, we identify two primary reasons for LLM inference inefficiency: 1) suboptimal GPU utilization due to lack of parallelism and memory-bound nature of decode phase, and 2) significant pipeline bubbles due to inconsistent prefill and decode times across different iterations, leading to micro-batch imbalance. To address these challenges, we introduce \sysname, a novel approach that incorporates \chunking and \hbatch. \Hbatch improves GPU utilization by piggybacking decodes with prefills, which converts the memory-bound decode phase to be compute bound. \Chunking helps with making more prefills available for decodes to piggyback, and also provides for a uniform unit of work which helps significantly reduce pipeline bubbles. We demonstrate that \sysname results in significant improvements in end-to-end throughput across models and hardware configurations.